\documentclass[conference]{IEEEtran}
\IEEEoverridecommandlockouts
\usepackage{cite}
\usepackage{amsmath,amssymb,amsfonts}
\usepackage{algorithmic}
\usepackage{graphicx}
\usepackage{textcomp}
\usepackage{xcolor}
\usepackage{array}
\usepackage{booktabs}
\usepackage{makecell}
\usepackage{subcaption}
\usepackage{hyperref}

\newcolumntype{C}[1]{>{\centering\arraybackslash}m{#1}} 
\newcolumntype{L}[1]{>{\arraybackslash}m{#1}}

\def\BibTeX{{\rm B\kern-.05em{\sc i\kern-.025em b}\kern-.08em
    T\kern-.1667em\lower.7ex\hbox{E}\kern-.125emX}}
\begin{document}

\def\dataset{TrueSkin}
\title{\dataset: Towards Fair and Accurate Skin Tone Recognition and Generation}

\author{\IEEEauthorblockN{1\textsuperscript{st}Haoming Lu}
\IEEEauthorblockA{\textit{Topaz Labs} \\
Dallas, United States \\
jszjlhm@gmail.com
}}

\maketitle

\begin{abstract}

Skin tone recognition and generation play important roles in model fairness, healthcare, and generative AI, yet they remain challenging due to the lack of comprehensive datasets and robust methodologies. Compared to other human image analysis tasks, state-of-the-art large multimodal models (LMMs) and image generation models struggle to recognize and synthesize skin tones accurately. To address this, we introduce TrueSkin, a dataset with 7299 images systematically categorized into 6 classes, collected under diverse lighting conditions, camera angles, and capture settings. Using TrueSkin, we benchmark existing recognition and generation approaches, revealing substantial biases: LMMs tend to misclassify intermediate skin tones as lighter ones, whereas generative models struggle to accurately produce specified skin tones when influenced by inherent biases from unrelated attributes in the prompts, such as hairstyle or environmental context. We further demonstrate that training a recognition model on TrueSkin improves classification accuracy by more than 20\% compared to LMMs and conventional approaches, and fine-tuning with TrueSkin significantly improves skin tone fidelity in image generation models. Our findings highlight the need for comprehensive datasets like TrueSkin, which not only serves as a benchmark for evaluating existing models but also provides a valuable training resource to enhance fairness and accuracy in skin tone recognition and generation tasks. The dataset is available for download \href{https://drive.google.com/file/d/1_ndw5uyY4h4DLL5iGTL4bVDKdE_g_H4B/view?usp=sharing}{here}.

\end{abstract}

\begin{IEEEkeywords}
skin tone recognition, model fairness, large multimodal model, generative model
\end{IEEEkeywords}

\section{Introduction}
\label{sec:intro}

\begin{figure}[!t]
  \centering  
   \includegraphics[width=\linewidth]{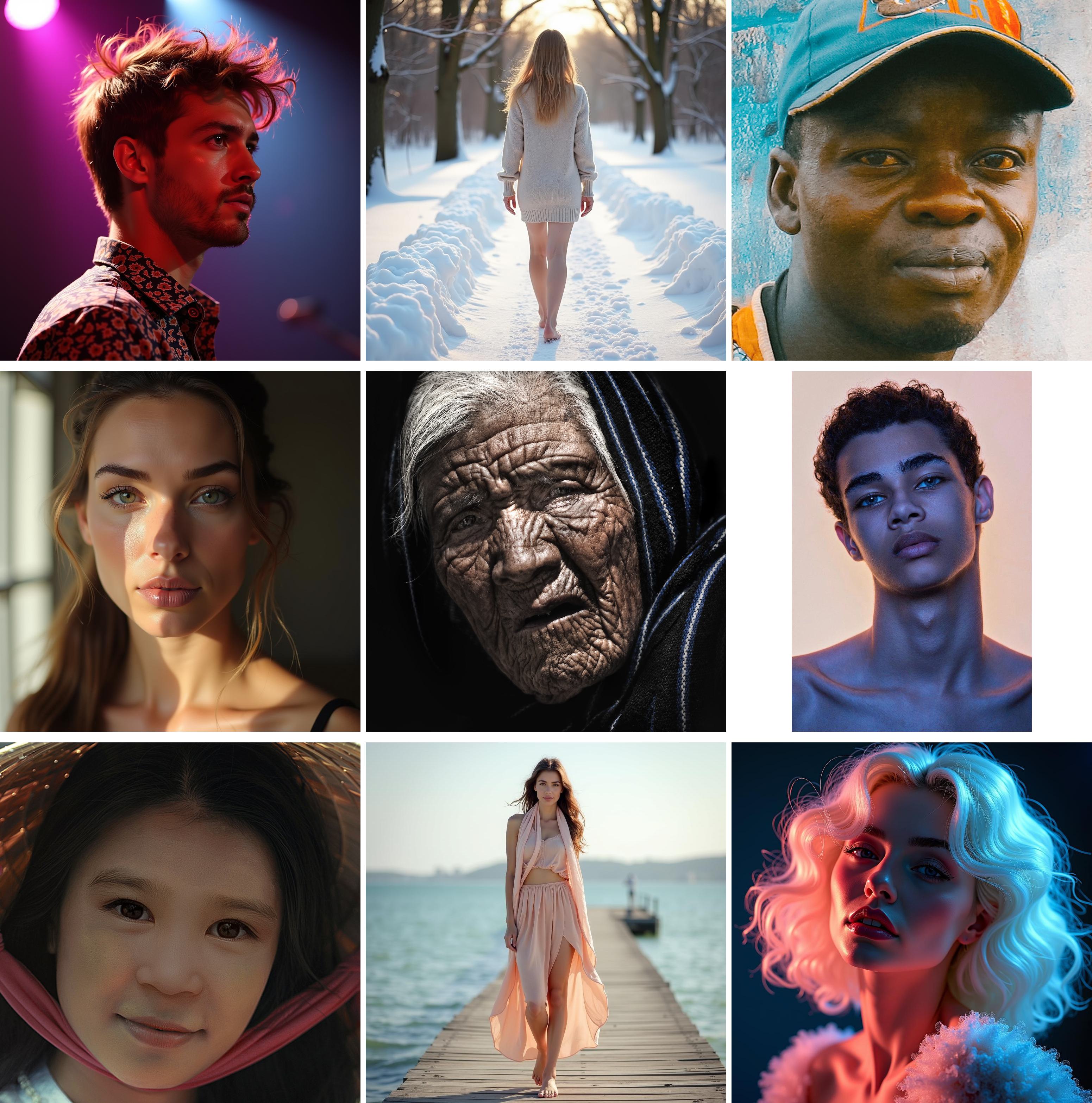}
   \caption{Samples from \dataset~dataset, showing the discrepancies between apparent and true skin tone. The labels of each image presented in left-to-right and top-to-bottom order are: medium, pale, dark, light, brown, tan, medium, light, and pale.}
   \label{fig:samples}
   \vspace{-10pt}
\end{figure}

Due to their widespread use in various applications, human images have long been a central focus of computer vision research. Studies in this domain span a wide range of tasks, such as face recognition~\cite{kim2022adaface}, pose estimation~\cite{zheng2023deep}, and person re-identification~\cite{ye2021deep}. However, compared to other tasks, skin tone analysis remains relatively underexplored despite its importance for model fairness~\cite{lohr2022facial, trueskinethic2, appcolor1}, healthcare~\cite{groh2024deep, cohen2023colorimetric, trueskinforfair1}, and generative models~\cite{bloomberg2023generative, chinchure2024tibet,lu2023specialist}. \par
For example, modern face encoders have achieved remarkable success, attaining over 99\% accuracy in face recognition~\cite{alansari2023ghostfacenets}, and effectively supporting identity-preserving image generation models through extracted face embeddings~\cite{wang2024instantid, guo2024pulid}. However, skin tone recognition and generation still face significant limitations. On one hand, state-of-the-art (SOTA) large multimodal models (LMMs) such as LLama~\cite{dubey2024llama}, LLaVA~\cite{liu2024llavanext}, and Phi~\cite{abdin2024phi}, encounter significant difficulties when recognizing skin tones in in-the-wild images with complex lighting and camera conditions. On the other hand, leading image generation models, such as SDXL~\cite{esser2024scaling}, SD3~\cite{esser2024scaling}, and FLUX.1~\cite{flux2024}, often fail to accurately generate the specified skin tone when influenced by inherent biases from unrelated prompt attributes, such as hairstyle or environmental context. These observations underscore the need for comprehensive datasets and improved methodologies for accurate skin tone recognition and synthesis. \par

\begin{figure}[t]
  \centering
  \begin{minipage}[t]{0.35\linewidth}
    \centering
    \includegraphics[width=\linewidth, height=5cm, keepaspectratio]{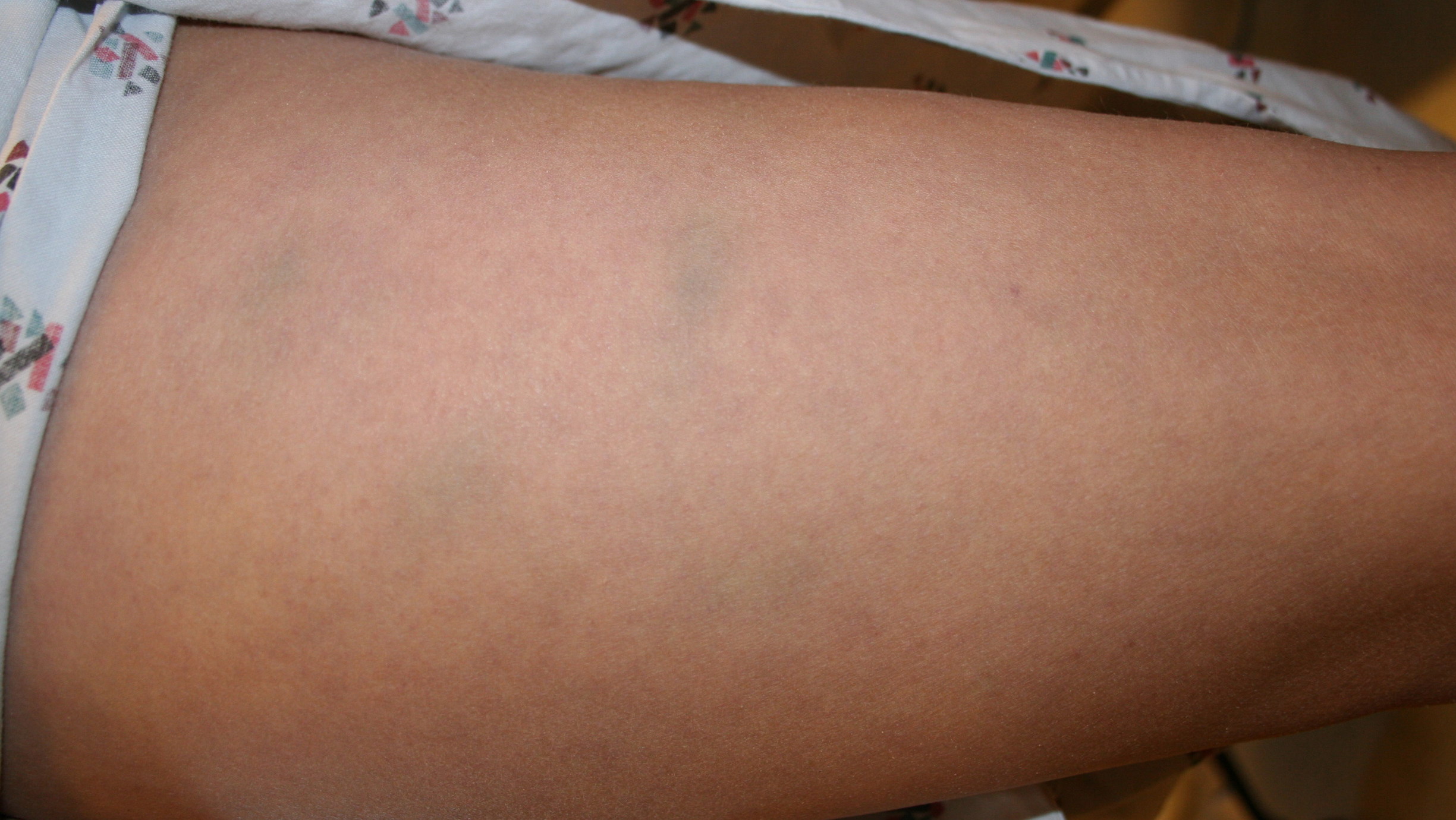}
  \end{minipage}
  \hfill
  \begin{minipage}[t]{0.58\linewidth}
    \centering
    \includegraphics[width=\linewidth, height=5cm, keepaspectratio]{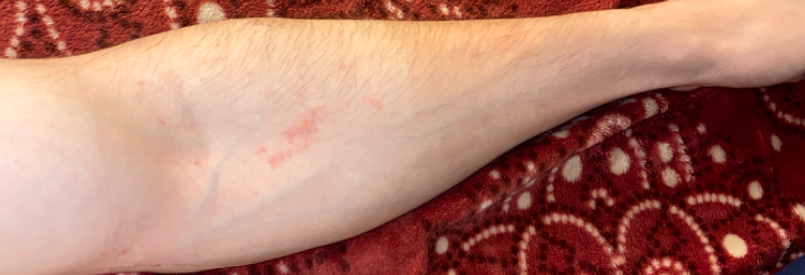}
  \end{minipage}
  \caption{Samples from existing skin tone datasets: Fitzpatrick17k (left) and SCIN (right). These datasets predominantly contain close-up images of isolated body parts, classified based on medical criteria (e.g., skin’s tendency to tan or burn), resulting in limited diversity in appearance and context.}  
  \label{fig:otherdata}
  \vspace{-10pt}
\end{figure}

The challenges in skin tone recognition and synthesis arise from multiple factors. First, due to variations in capture conditions, such as lighting position and color, the apparent skin tone in an image may significantly deviate from an individual's actual skin tone. Existing methods aimed at mitigating these discrepancies~\cite{krishnapriya2022analysis, lester2021clinical} are often tailored to specific datasets or capture settings, making them difficult to generalize. Second, compared to attributes like ethnicity and gender, human image datasets with explicit skin tone annotations are severely limited. Existing skin tone datasets~\cite{scin, fit17k} are predominantly derived from medical environments. As illustrated in Fig. \ref{fig:otherdata}, they consist mainly of close-up shots of specific body parts, and their classification criteria take into account medical considerations, such as whether the skin tans or burns under ultraviolet radiation. The focus of these datasets differs from the primary interests of computer vision research,  limiting their contribution to training skin tone recognition or generation models that cater to general applications. Finally, skin tone is inherently subjective to some extent~\cite{barrett2023skin}, as perceptions of skin tone can vary across individuals. This subjectivity makes establishing accurate and consistent annotations significantly more challenging than that for objective attributes such as identity or age. \par

In conclusion, the primary limitation hindering the performance of existing methods in skin tone recognition and generation is the lack of a dataset that contains sufficient diversity and precise annotations. To address this gap, we introduce \textbf{\dataset} dataset, which comprises 7,299 images collected from diverse sources and spanning varied lighting conditions, camera angles, and capture settings. The dataset is systematically categorized into six distinct skin tone classes based on a clear and interpretable classification standard. \dataset~serves as both a benchmark for evaluating the performance of existing approaches and a training resource to improve models for skin tone-related tasks.

The remainder of this paper is structured as follows: Sec. \ref{sec:background} discusses the impact of skin tone recognition and generation and highlights the limitations of existing datasets and methods. Sec. \ref{sec:dataset} details the construction of \dataset~dataset and outlines its improvements over prior datasets. In Sec. \ref{sec:vlm}, we use \dataset~to evaluate the performance of SOTA LMMs and conventional skin tone recognition approaches, and analyze their respective error patterns. Sec. \ref{sec:diffusion} assesses the ability of leading image generation models to produce images with specified skin tones and examines the potential sources of bias. In Sec. \ref{sec:baseline}, we train a baseline recognition model using TrueSkin as a training dataset, demonstrating its effectiveness in improving the performance of skin tone recognition. Sec. \ref{sec:finetune} investigates the fine-tuning of existing image generation models using \dataset~and evaluates its capability in reducing model bias. Finally, Sec. \ref{sec:conclusion} concludes the experiments, discusses the limitations of \dataset, and outlines potential directions for future research. The contributions of this paper can be summarized as follows:

\begin{enumerate}

\item We introduce \dataset, a high-quality skin tone dataset that incorporates diverse lighting conditions, camera angles, and data sources, systematically categorized into six distinct skin tone classes. The dataset provides a benchmark for evaluating existing skin tone recognition and generation methods and serves as a training resource to improve model performance.

\item We systematically analyze the performance and biases of state-of-the-art large multimodal models and image generation models on skin tone recognition and synthesis tasks using \dataset. Our study reveals significant limitations in both classification accuracy and generative consistency of existing models.

\item To address these limitations, we leverage \dataset~for both recognition and generation model training. We demonstrate that fine-tuning existing image generation models with \dataset~helps mitigate bias, while training a simple recognition model on \dataset~significantly improves classification accuracy over existing methods.  

\end{enumerate}

\section{Background}
\label{sec:background}

True skin tone refers to an individual's inherent skin tone independent of external influences such as lighting, camera settings, or image processing, whereas apparent skin tone can fluctuate under different environmental conditions. For instance, photos of the same person taken in bright sunlight and in dim settings may display different skin tones despite the person's actual skin color being consistent. Consider the portrait at the bottom-right of Fig. \ref{fig:samples}: Due to the strong red lighting covering most of her visible skin, the apparent skin tone extracted from the skin pixels appears overly red. However, by examining the small low-light region in the center of her face, it can be inferred that her true skin tone is pale. Furthermore, the true skin tone is also distinct from ethnicity, which embodies cultural and ancestral identity rather than a mere physical trait. As noted in \cite{skin-and-ethic}, individuals within the same ethnic group may exhibit a range of skin tones, and similar tones can be observed across different ethnicities. \par

True skin tone plays an important role in both computer vision and medical research. In computer vision, face recognition models show reduced accuracy for individuals with darker skin tones~\cite{lohr2022facial, trueskinethic2}, while gender classification and image cropping algorithms demonstrate biases favoring specific skin tones~\cite{appcolor1}. Such bias is also evident in generative models~\cite{bloomberg2023generative, chinchure2024tibet}, as seen in their tendency to associate darker skin tones with lower-income professions. In medicine, precise skin tone identification is critical for dermatology and cosmetology, where they support diagnosis, treatment, and personalized care~\cite{cohen2023colorimetric, trueskinforfair1, trueskinforfair2, trueskinforfair3, trueskinforfair4}. Additionally, evidence suggests that pulse oximetry devices may exhibit varying accuracy across different skin tones, potentially increasing the risk of undetected hypoxemia in individuals with darker skin~\cite{sjoding2020racial, adler1998effect, ebmeier2018two}. Furthermore, studies have shown that physicians exhibit reduced diagnostic accuracy when diagnosing patients with darker skin tones~\cite{groh2024deep}. These findings highlight how the failure to recognize or process diverse skin tones can result in significant biases and adverse outcomes. \par

A variety of approaches have been explored to address these challenges. First, to achieve a more precise definition of skin tones, numerous scales have been proposed, such as the Fitzpatrick scale~\cite{fitzpatrick1975}, the Monk scale~\cite{monk2019monk}, and the von Luschan scale~\cite{swiatoniowski2013comparing}. However, since fine-grained scales increase annotation complexity, and skin tone labeling is susceptible to biases introduced by annotators' backgrounds~\cite{barrett2023skin, lu2024can}, there is no consensus on the optimal scale in practice yet. Second, since apparent skin tone can be readily obtained through skin pixel segmentation~\cite{shaik2015comparative}, some studies~\cite{krishnapriya2022analysis, lester2021clinical} have attempted to infer true skin tone by mitigating the influence of environmental factors such as lighting conditions and camera settings. Nonetheless, these approaches have limited scope of application and do not adequately account for scenarios with colored or multiple light sources. Furthermore, some recent medical image datasets~\cite{groh2022towards, daneshjou2022disparities} provide a more balanced distribution of skin tones. However, these datasets are often limited in size, and medical research tends to focus on reducing the discrepancies between true and apparent skin tone during image acquisition rather than analyzing their differences~\cite{weir2024survey}. As a result, their applicability to true skin tone research remains limited. On the other hand, while many human image datasets~\cite{fairface, utkface, celeba, racialfaces} include ethnicity as an attribute, the few existing datasets with skin tone annotations suffer from issues such as insufficient sample size~\cite{schumann2023consensus} or overly coarse categories~\cite{rana2023skin}. \par

In conclusion, true skin tone differs notably from apparent skin tone and ethnicity. Despite its significant impact on computer vision and medical research, true skin tone remains insufficiently studied due to limited datasets. Therefore, the development of new datasets and tools to advance research in this area is both essential and urgent~\cite{harvey2024integrating}.

\section{TrueSkin dataset}
\label{sec:dataset}

\begin{figure*}[tb]
   \centering  
   \includegraphics[width=0.9\linewidth]{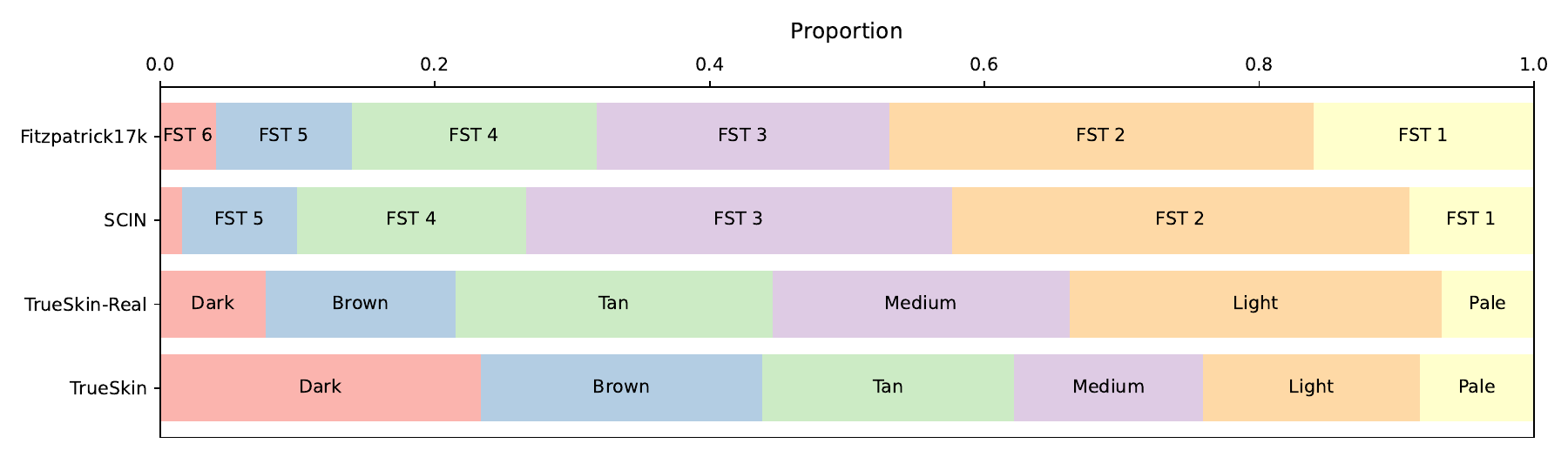}
   \caption{Skin tone label distribution of existing datasets and \dataset.}
   \label{fig:datasetdist}
   \vspace{-12pt}
\end{figure*}

\subsection{Dataset construction}
As previously discussed, building a high-quality skin tone dataset requires addressing two key challenges: subjectivity and bias. To address these challenges, we developed~\dataset~with carefully defined criteria. In concrete terms, the dataset is constructed based on the following principles:

\paragraph{Consistency}

\begin{table}[tb]
\centering
\caption{Classification standards for the six skin tones in natural language. The example image set expands during the annotation process to reduce discrepancies.}
\label{tab:anno}
\begin{tabular}{C{1.2cm}|L{3.8cm}|C{2cm}}
\toprule
Label & Description & Example \\
\midrule
Dark & Deepest skin tone, from deep brown to nearly black, stays dark in sunlight, and appears almost black in low light. & 
\includegraphics[width=0.8\linewidth]{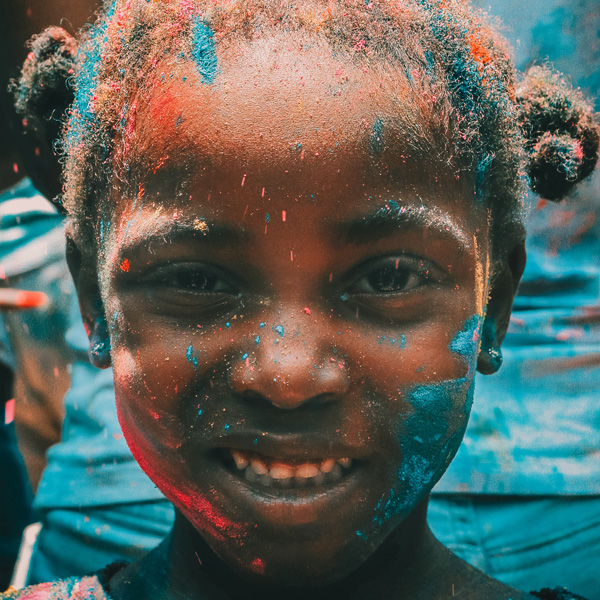} \\
\midrule
Brown & Deep skin tone but not extremely dark, visibly brown even in low light conditions. & 
\includegraphics[width=0.8\linewidth]{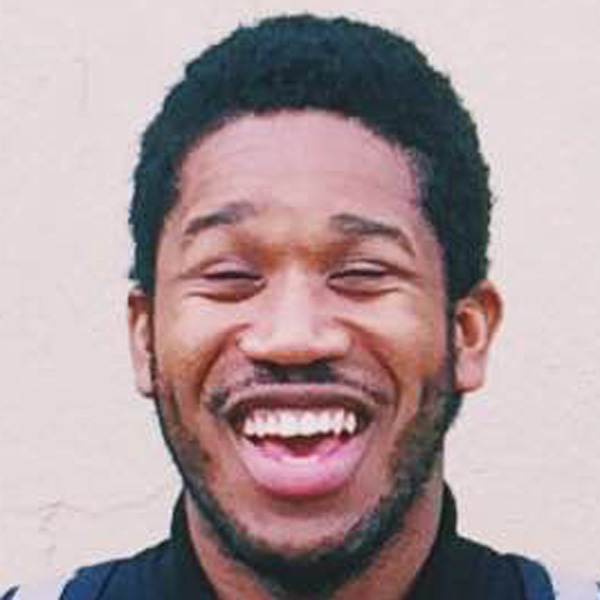} \\
\midrule
Tan & Lighter brown, appears golden in sunlight, looks muted, and often shifts to soft brown in low light.  & 
\includegraphics[width=0.8\linewidth]{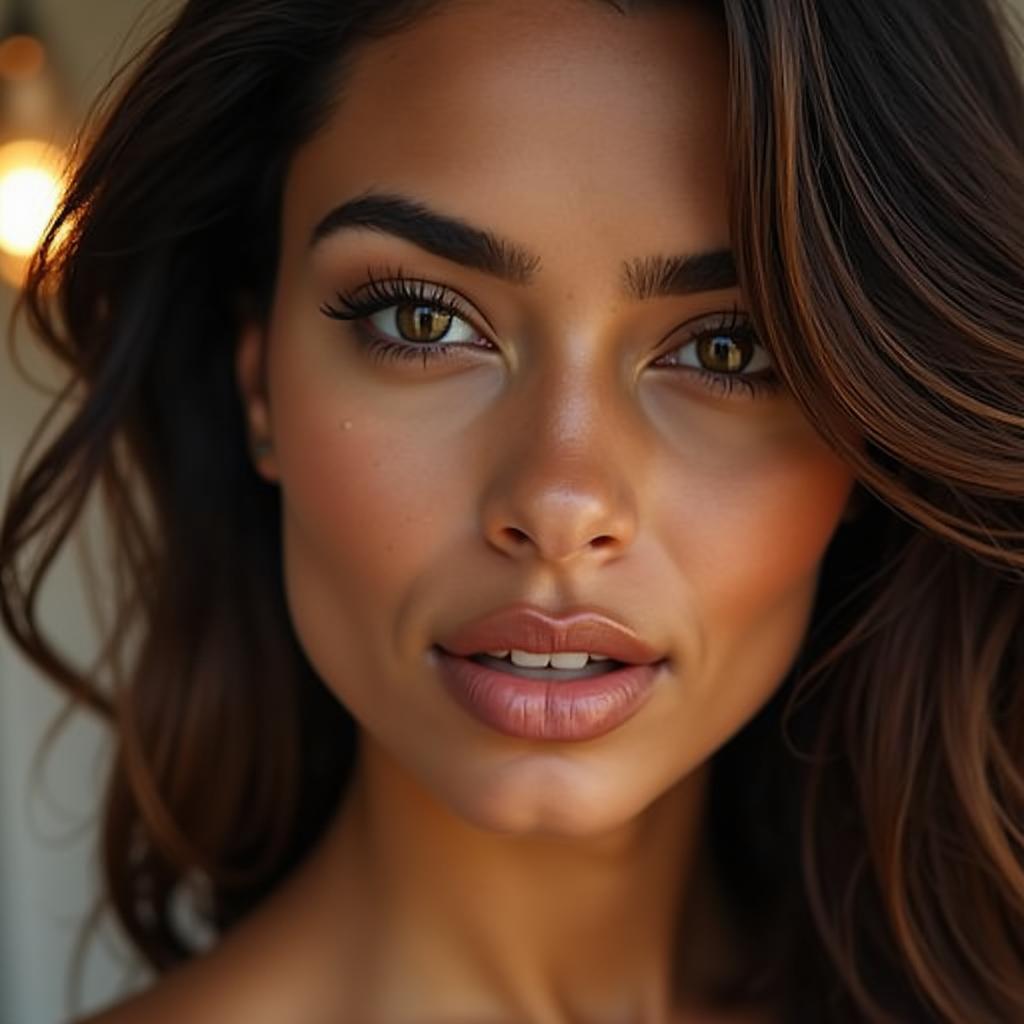} \\
\midrule
Medium & Neutral and balanced tone, retains more color depth than tan skin in low light & 
\includegraphics[width=0.8\linewidth]{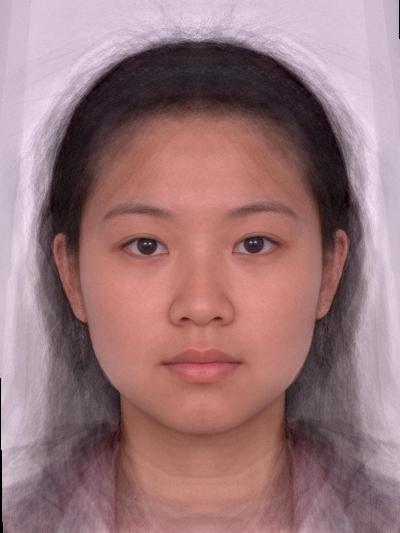} \\
\midrule
Light & Appears creamy to fair in sunlight, holds warmth and visibility in low light. & 
\includegraphics[width=0.8\linewidth]{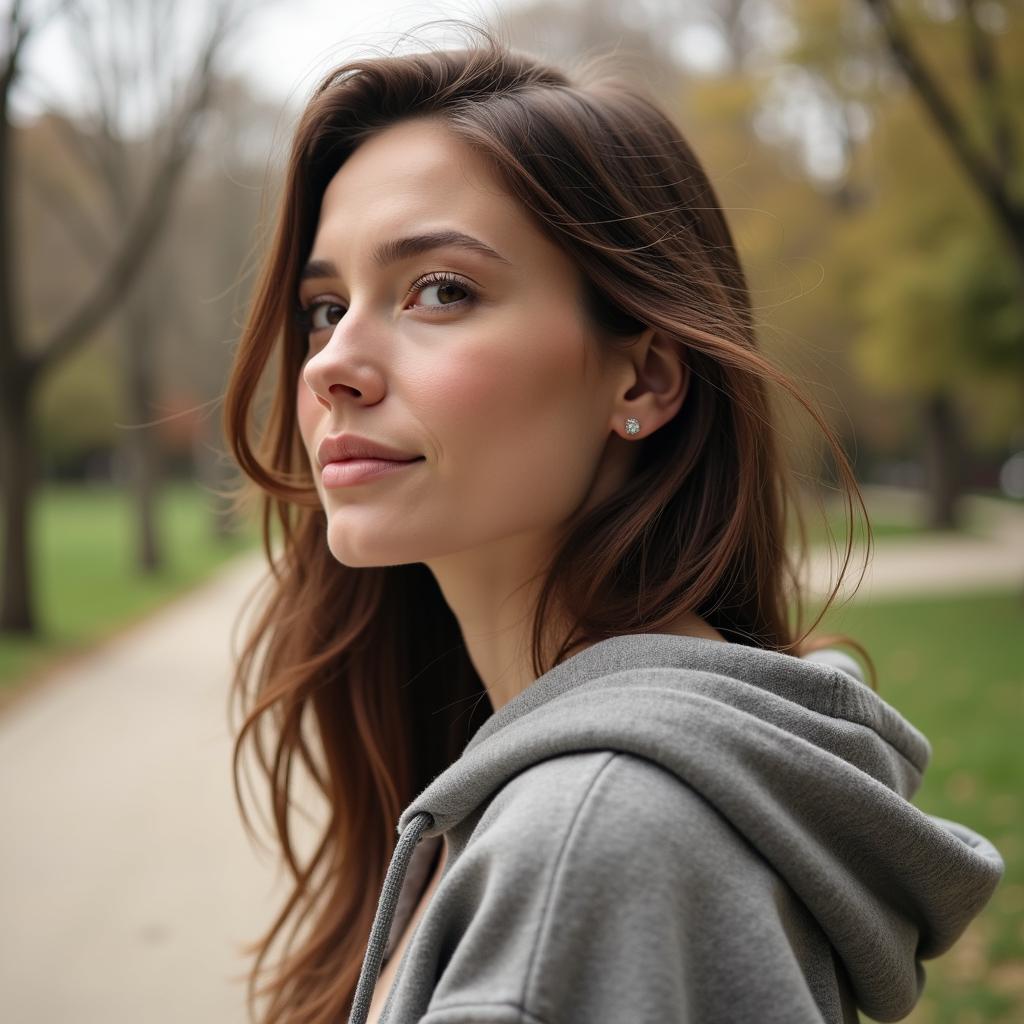} \\
\midrule
Pale & Lightest skin, leans cooler than light skin, appears almost colorless in low light. & 
\includegraphics[width=0.8\linewidth]{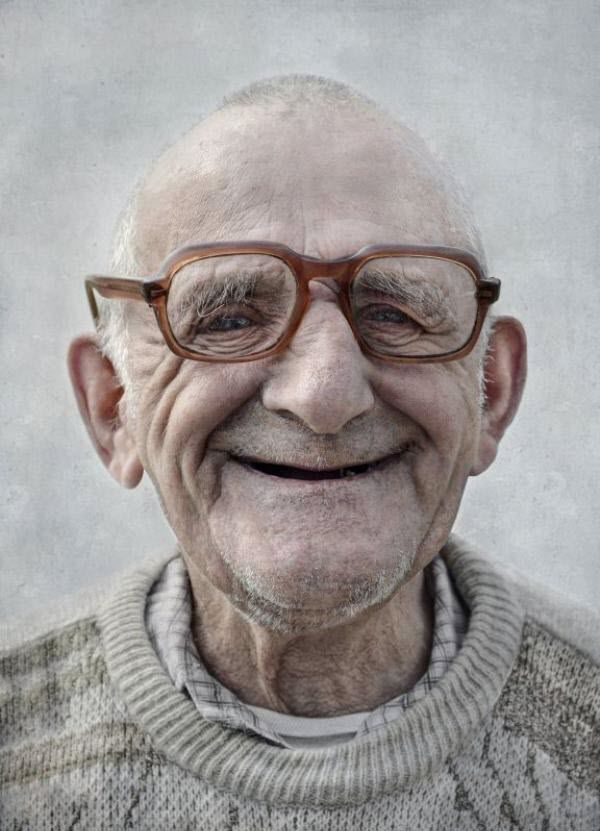} \\
\bottomrule
\end{tabular}
\vspace{-10pt}
\end{table}

Like any dataset, ensuring the correctness of annotations is paramount for~\dataset. The Fitzpatrick scale~\cite{fitzpatrick1975}, which was originally designed to classify skin types by their reaction to ultraviolet light (for example, FST 1 refers to people who always burn and never tan), introduces divergence and bias when applied through visual assessment alone, as shown in~\cite{ware2020racial, bhanot2024fitzpatrick}. Conversely, the Monk~\cite{monk2019monk} and von Luschan~\cite{swiatoniowski2013comparing} scales, offering 10 and 36 categories respectively, are excessively fine-grained for practical visual use. We attempted to annotate \dataset~using these fine-grained scales with the method described later; however, less than 30\% of the samples achieved valid consensus. After careful consideration, we decided to adopt the six-category structure of the Fitzpatrick scale, but with a classification standard based solely on visual perception rather than medical criteria, thereby reducing cognitive complexity and potential discrepancies. To facilitate consistent labeling, each category is assigned a distinct name (dark, brown, tan, medium, light, pale) and supplemented with representative sample images and detailed descriptions. As shown in Table. \ref{tab:anno}, unlike existing medical datasets, our classification of skin tones is based purely on visual perception. The annotation was conducted by six annotators from diverse ethnic backgrounds, and only samples for which at least four annotators reached a consensus were included in the dataset. These measures ensure internal consistency and reduce subjectivity in annotation, enhancing the overall reliability of the dataset.

\paragraph{Balanced}
The performance disparities of existing models across different skin tones are largely attributed to the imbalanced distribution of samples in datasets. To mitigate this issue, we aim to ensure that the final dataset maintains a more uniform sample distribution across categories. Fortunately, although Sec. \ref{sec:diffusion} will demonstrate that image generation models do not always produce images with the intended skin tones, they can still be leveraged to supplement underrepresented categories, improving the overall balance within the dataset.

\begin{table}[tb]
  \centering
  \caption{Coefficient of Variation (CV) and Kullback-Leibler (KL) divergence of datasets. In both metrics, lower values indicate a more uniform distribution.}
  \label{tab:unbias}
  \begin{tabular}{lcc}
    \toprule
    Dataset\textbackslash Metrics & CV~\textdownarrow & KL Divergence~\textdownarrow \\
    \midrule
    Fitzpatrick17k & 0.5566 & 0.2013 \\
    SCIN & 0.7750 & 0.3926 \\
    \dataset~(real only) & 0.5106 & 0.1680 \\
    \dataset & \textbf{0.3179} & \textbf{0.0652} \\     
    \bottomrule
  \end{tabular}
\end{table}

\paragraph{Diversity}
To enhance the robustness of skin tone recognition or generation models, we introduce multiple dimensions of diversity in ~\dataset. First, it includes images captured under various lighting conditions, differing in color, intensity, and angle. Second, in medical datasets, images are typically close-ups of the skin, where the high proportion of skin pixels results in minimal environmental context. This limitation prevents models from effectively learning to disentangle skin tone from lighting influences, forcing them to rely on prior knowledge or dataset-specific biases. \dataset~includes a diverse range of framing compositions, including close-ups, full-body shots, and non-facial images with varying skin pixel proportions, allowing models trained on it to infer true skin tone across different conditions. Finally, since skin condition can evolve with age, with older individuals more likely to develop wrinkles and dark spots, failure to account for these variations can impact model performance. Therefore, we incorporate a broad range of ages in ~\dataset~ from infants to the elderly. The selected samples in Fig. \ref{fig:samples} visually illustrate the mentioned aspects of diversity.

\subsection{Dataset Statistics}

\begin{table}[tb]
  \setlength{\tabcolsep}{5pt}
  \caption{Distribution of skin tone categories in \dataset}  
  \label{tab:distcount}
  \centering
  \begin{tabular}{ccccccc}
    \toprule
    & Dark & Brown & Tan & Medium & Light & Pale \\
    \midrule
    Real & 138 & 247 & 413 & 387 & 485 & 120 \\
    Gen & 1563 & 1251 & 923 & 617 & 667 & 488 \\
    \midrule
    Total & 1701 & 1498	& 1336 & 1004 & 1152 & 608 \\
    \bottomrule
  \end{tabular}
  \vspace{-12pt}
\end{table}

\begin{figure}[tb]
  \centering  
   \includegraphics[width=\linewidth]{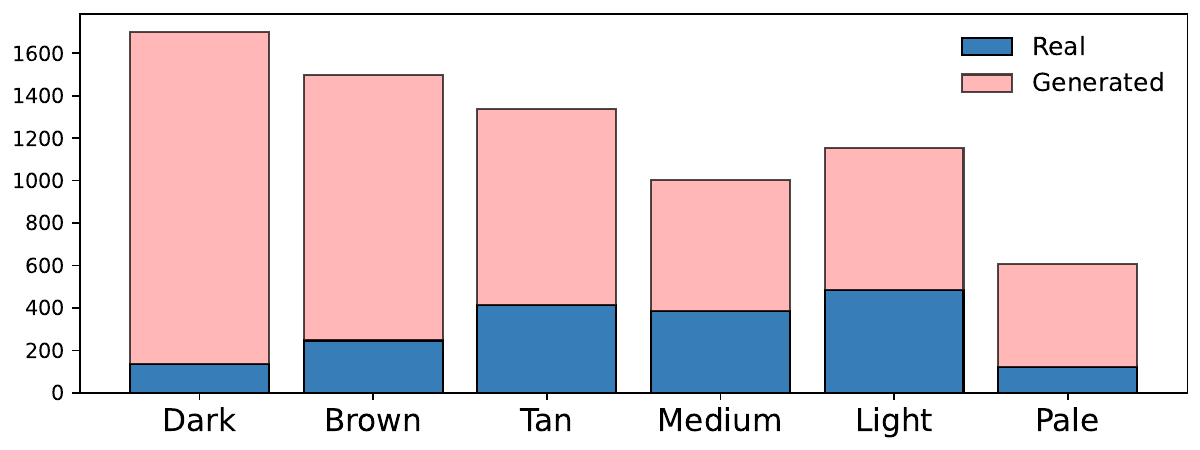}
   \caption{Count of real images versus generated images.}   
   \label{fig:distbar}
   \vspace{-12pt}
\end{figure}

\begin{figure}[tb]    
    \centering
    \includegraphics[width=0.65\linewidth]{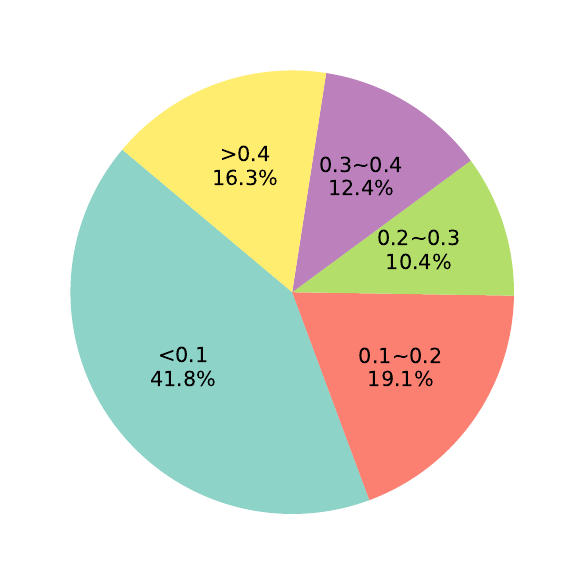}    
    \caption{Distribution of the proportion of skin area relative to total image area in the dataset. $<0.1$ means less than 10\% of the pixels are skin areas.}
    \label{fig:humanratio}
    \vspace{-12pt}
\end{figure}

\begin{figure}[tb]
  \centering  
    \includegraphics[width=0.75\linewidth]{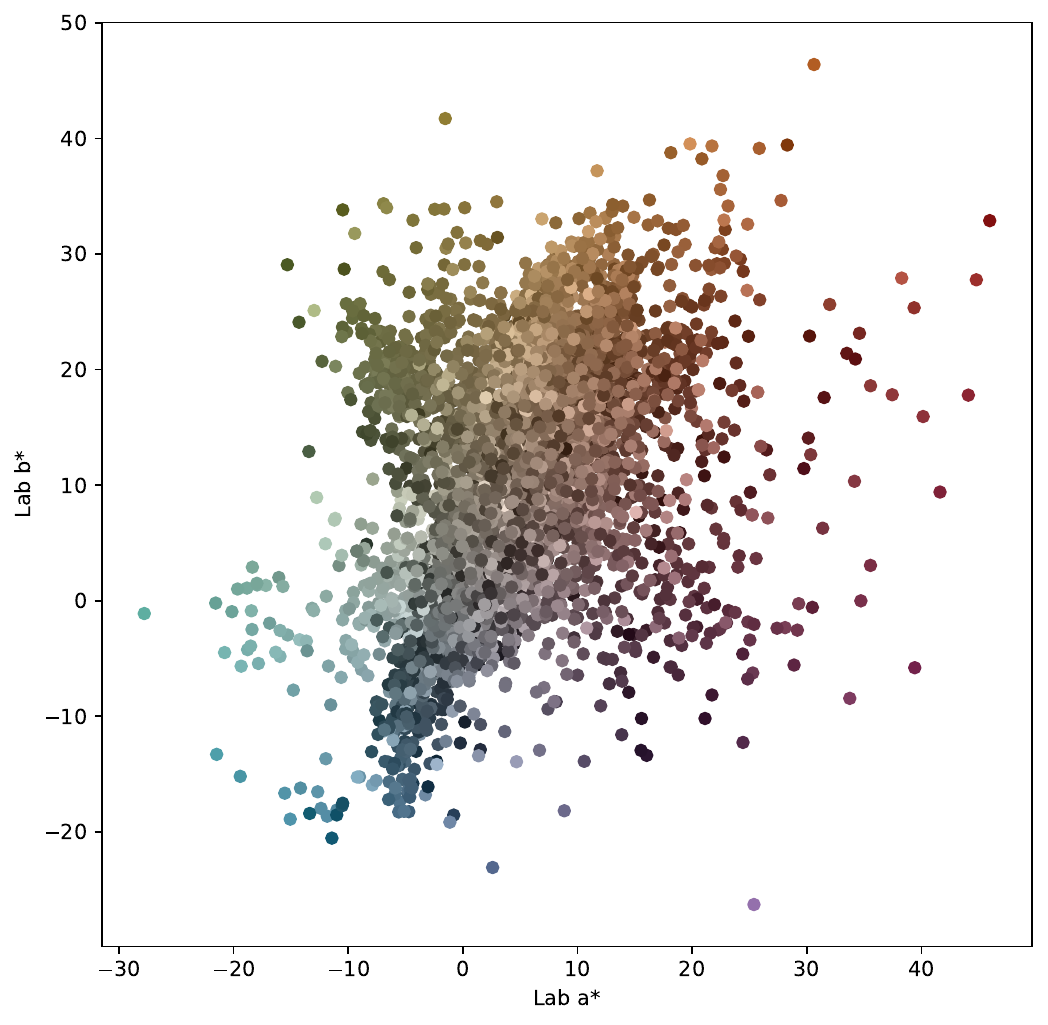}  
  \caption{Average color of each image in $Lab$ space, with the x-axis ($a$) from greenish to reddish and the y-axis ($b$) from bluish to yellowish.}
  \label{fig:diversity}
  \vspace{-18pt}
\end{figure}

The detailed composition of \dataset~is presented in Table. \ref{tab:distcount}. The real images are sourced from~\cite{realdata}, while the synthetic images are generated using the FLUX.1-dev. As demonstrated in Fig. \ref{fig:distbar}, incorporating synthetic data reduces the concentration of samples in the middle skin tones, leading to a more balanced distribution across categories. We compare the label distribution of TrueSkin with existing skin tone datasets. Fig. \ref{fig:datasetdist} provides a visual representation of this comparison, while Table. \ref{tab:unbias} quantifies the differences in label distribution. On the other hand, as shown in Fig. \ref{fig:humanratio} and Fig. \ref{fig:diversity}, \dataset~exhibits substantial diversity in both skin pixel proportions and overall image color composition, which reflects the variance in lighting conditions.

\section{Challenges with Large Multimodal Models}
\label{sec:vlm}

\subsection{Models and Experiment Setup}
To comprehensively evaluate the ability of current LMMs in skin color recognition, we selected a diverse set of leading open-source models: LLaMA 3.2~\cite{dubey2024llama}, LLaVA~\cite{liu2024llavanext}, Janus-Pro~\cite{chen2025janus}, Qwen 2.5-VL~\cite{qwen2.5-VL}, and Phi 3.5~\cite{abdin2024phi}. Additionally, as a baseline, we use a conventional method that segments the skin region, clusters pixel-level colors, computes the ITA of the largest cluster’s centroid, and maps it to one of six categories following \cite{fit17k}. \par

\begin{table}[b]
  \vspace{-10pt}
  \caption{Mapping from LMM outputs to dataset labels}
  \label{tab:vlm-mapping}
  \centering
  \begin{tabular}{ll}
    \toprule
    Dataset label & LMM outputs \\
    \midrule
    Dark & Dark, Black \\
    Brown & Brown \\
    Tan & Tan, Tanned \\
    Medium & Blond, Blonde, Beige, Yellow \\
    Light & Fair, Light, White, Caucasian \\
    Pale & Gray, Grey, Pale, Albino \\
    \bottomrule
  \end{tabular}
\end{table}

Using LMMs for skin tone recognition is not straightforward, as they tend to produce lengthy, ambiguous responses rather than direct answers. Recognizing that converting verbose LMM outputs into definitive skin tone labels introduces additional ambiguity, we opted for prompting approaches that elicit more structured outputs. Particularly, we explored three prompting strategies: (1) requesting a skin tone value in the RGB/Lab color space, (2) asking for one of the six dataset labels, either with or without providing their definitions, and (3) asking for a single-word description, which is then mapped to the dataset labels. The first approach fails in most cases, while the second and third achieve similar performance, with the third showing a slight advantage. Therefore, we adopt the third method. Specifically, we use the following prompt and filter out invalid responses such as ``old'', ``blue'', or ``unanswerable''. The remaining responses are mapped to dataset labels as shown in Table. \ref{tab:vlm-mapping}.

\begin{quotation}
\textit{You are given an image of a person. Determine the person's skin tone and use only one word as your response without any additional explanation.}
\end{quotation}

\subsection{Results and Observations}
\begin{figure*}[tb]
  \centering  
   \includegraphics[width=0.95\linewidth]{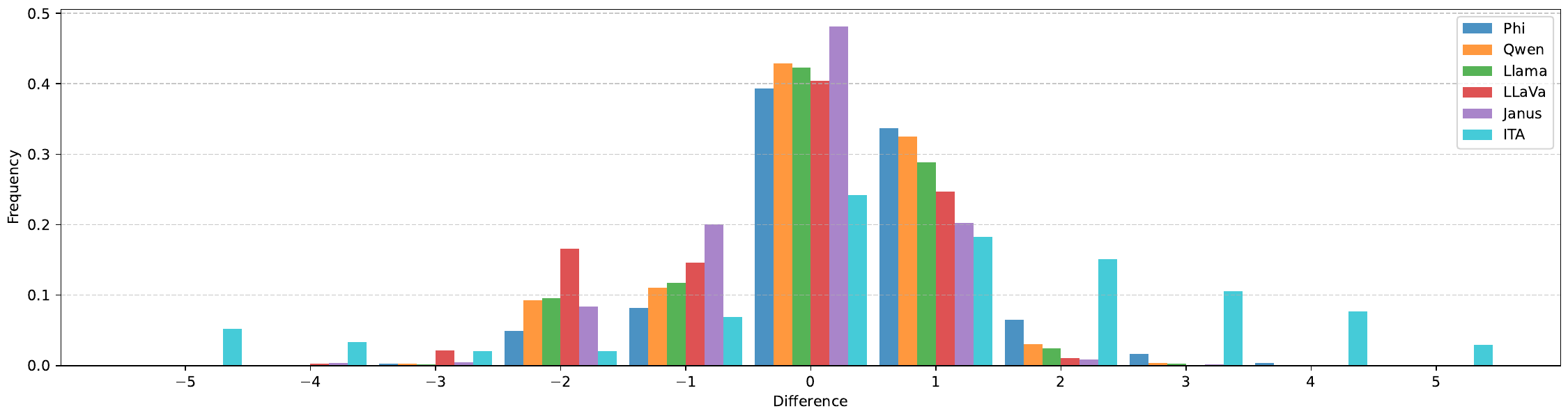}
   \vspace{-6pt}
   \caption{Overview of LLM evaluation: The x-axis represents the difference between the predicted and actual skin color indices, where a positive value indicates the predicted color is lighter than the true color. For example, a value of +2 corresponds to samples where the predicted index is 2 higher than the actual index, i.e. the predicted color is two shades lighter.}   
   \label{fig:vlmoverview}
   \vspace{-12pt}
\end{figure*}

Since skin tone recognition requires both accuracy and minimal deviation in misclassifications, we map the 6 labels to integers from 0 to 5 (dark = 0, pale = 5) to quantify the prediction error. Fig. \ref{fig:vlmoverview} and Table. \ref{tab:vlm-stat} present the overall performance of large multimodal models on \dataset~dataset. Most models achieve an accuracy between 0.4 and 0.5, with a bias toward lighter skin tones. Meanwhile, the traditional method performs significantly worse. \par

\begin{figure}[bt]
\centering
\renewcommand{\arraystretch}{1} 
\setlength{\tabcolsep}{4pt} 
\begin{tabular}{l c c}
    \toprule
    & \includegraphics[width=0.35\linewidth]{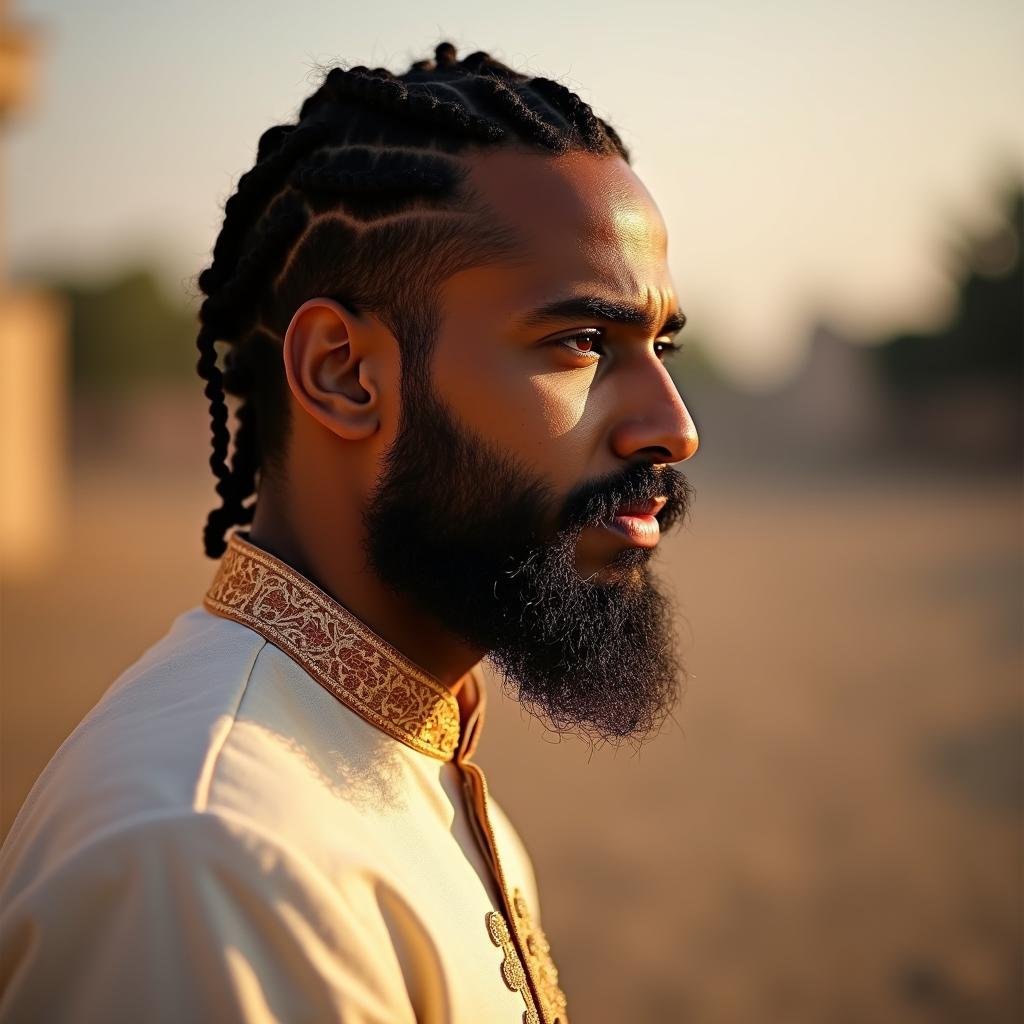} & \includegraphics[width=0.35\linewidth]{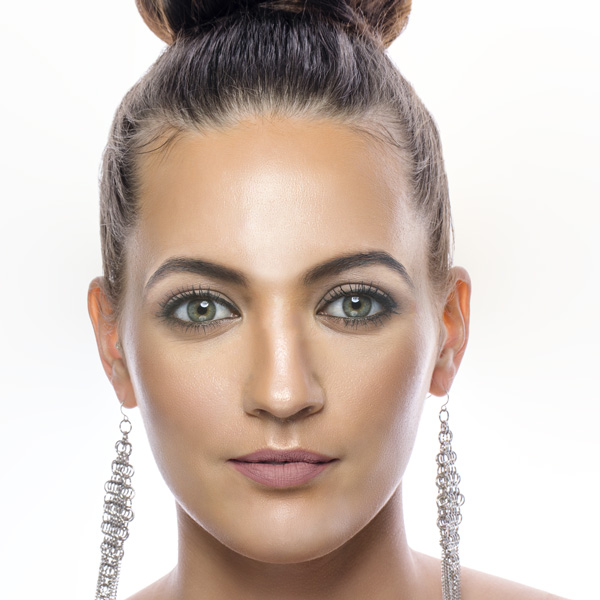} \\    
    \midrule
    LLaMA  & Dark   & Medium  \\
    LLaVA  & Dark   & Light   \\
    Janus  & Dark   & Light   \\
    Qwen   & Dark   & Light   \\
    Phi    & Dark   & Light   \\
    ITA    & Dark   & Light   \\
    \midrule
    \textbf{Truth} & \textbf{Brown} & \textbf{Tan} \\
    \bottomrule
\end{tabular}
\caption{Examples from \dataset~where all recognition approaches make incorrect predictions: In the left, most of the skin is backlit, appearing darker, but the contrast between high- and low-light areas indicates brown rather than dark. In the right, strong reflections on the face create a light appearance, while the lower-lit edges more accurately reflect the tan skin.}
\label{fig:vlm-example}
\vspace{-10pt}
\end{figure}

\begin{table}[tb]
\caption{Comparison of true skin recognition performance among LMMs and the baseline trained on~\dataset.}  
  \label{tab:vlm-stat}
  \centering
  \vspace{-4pt}
  \begin{tabular}{lcc}
    \toprule
    Model & Accuracy & MSE \\
    \midrule
    Llama3.2 & 44.31\% & 0.9866 \\    
    LLaVa-NeXT & 40.45\% & 1.3501 \\
    Janus-Pro-7B & 48.83\% & 0.8880 \\
    Qwen2.5 & 43.12\% & 1.0135 \\
    Phi-3.5 & 41.40\% & 1.1895 \\
    ITA & 24.84\% & 5.8816 \\
    \midrule
    \dataset & 74.18\% & 0.3374 \\
    \bottomrule
  \end{tabular}
\end{table}

\begin{figure*}[tb]
  \centering
  \begin{subfigure}{0.32\linewidth}
    \includegraphics[width=\linewidth]{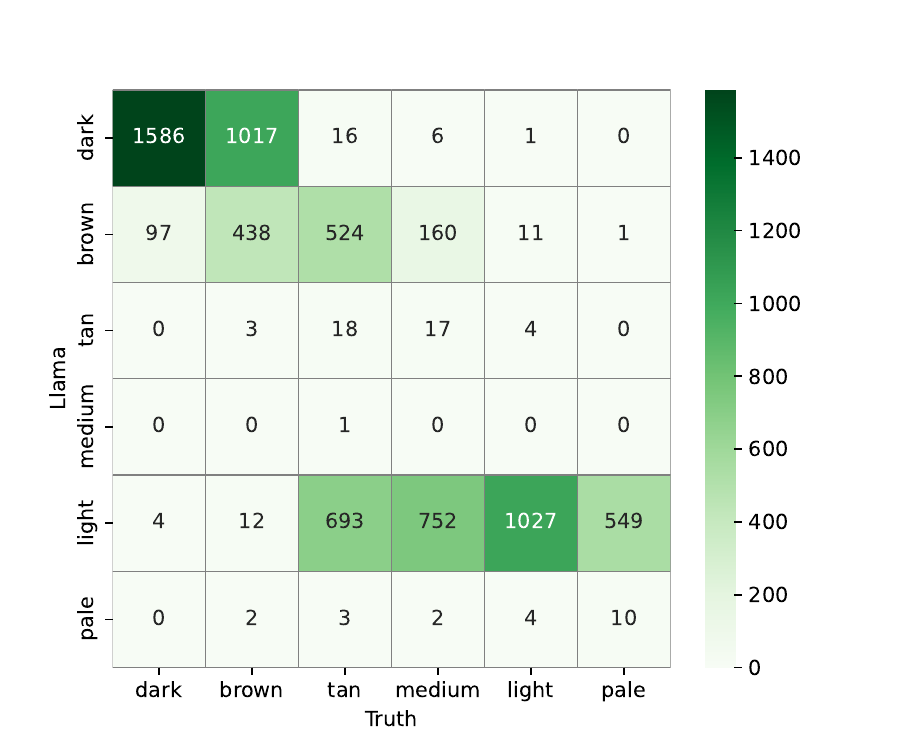}
    \caption{Llama-3.2-11B-Vision-Instruct}
    \label{fig:heat-llama}
  \end{subfigure}
  \hfill
  \begin{subfigure}{0.32\linewidth}
    \includegraphics[width=\linewidth]{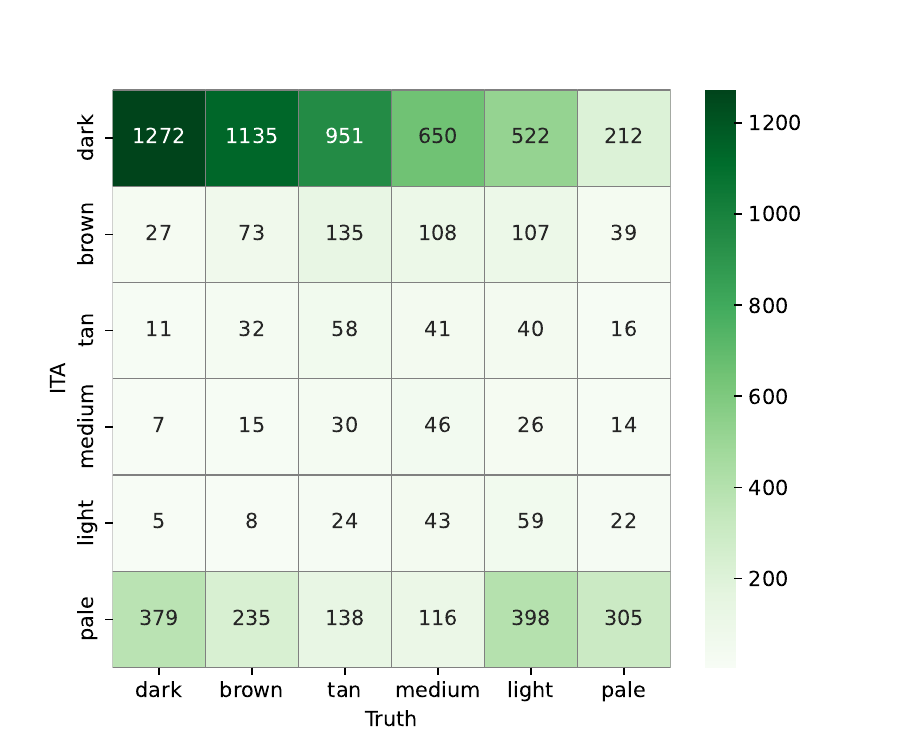}
    \caption{Traditional method (ITA)}
    \label{fig:heat-ita}
  \end{subfigure}
  \hfill
  \begin{subfigure}{0.32\linewidth}
    \includegraphics[width=\linewidth]{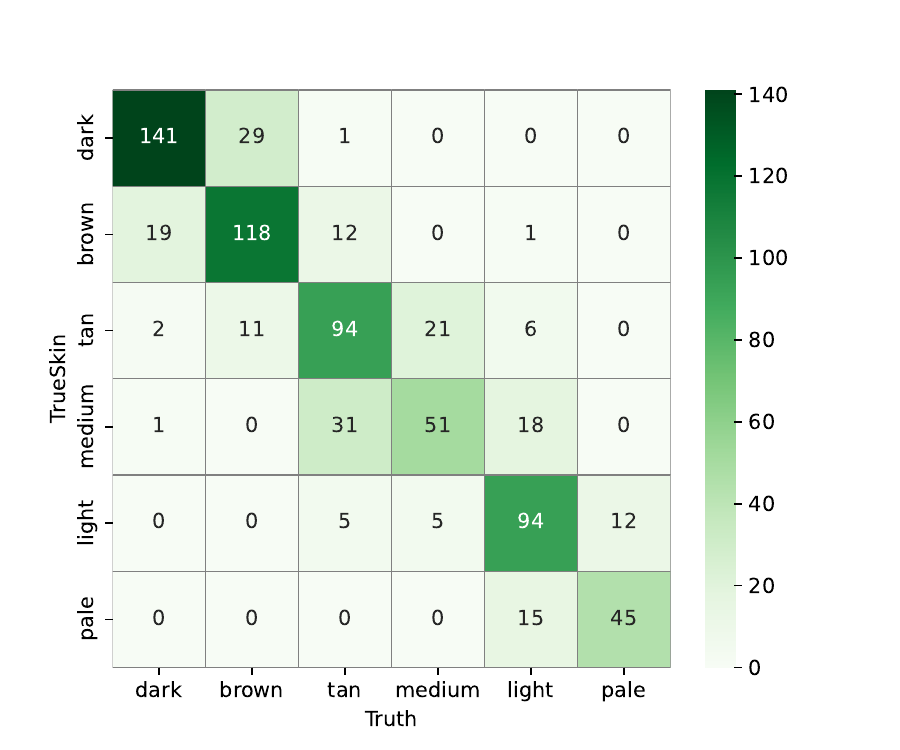}
    \caption{\dataset~baseline}
    \label{fig:heat-trueskin}
  \end{subfigure}
  \caption{Confusion matrices for different skin tone recognition approaches. Llama and ITA are evaluated on the entire dataset, while \dataset~baseline is assessed on the test set only.}  
  \label{fig:short}
  \vspace{-18pt}
\end{figure*}

While different LMMs show slight variations, their overall error patterns are similar. As shown in LLaMA’s confusion matrix in Fig.~\ref{fig:heat-llama}, the main issues are: (1) many brown samples misclassified as dark, and (2) the model failing to recognize the two middle skin tones and pale, often assigning them to brown or light. The examples in Fig. \ref{fig:vlm-example} illustrate how these errors may occur. On the other hand, the traditional pixel clustering approach essentially computes the apparent skin color. Thus, the confusion matrix in Fig. \ref{fig:heat-ita} reflects the discrepancy between apparent and true skin tones in \dataset. Due to the intentionally diverse lighting conditions, the difference is particularly pronounced. 

\section{Challenges with Image Generation Models}

\label{sec:diffusion}

\subsection{Models and Experiment Setup}
To evaluate image generation models' capability in rendering specified skin tones, we selected three leading open-source models: SDXL~\cite{podell2023sdxl}, SD3-Large-Turbo~\cite{esser2024scaling}, and FLUX.1 Dev~\cite{flux2024}. To better reflect real-world scenarios, we generate prompts with ChatGPT-4o~\cite{OpenAI2024GPT4o} that explicitly included a skin tone description in the format ``[target tone] skin'' while varying other factors like hairstyle, lighting, clothing, background, and camera distance. For each skin tone and gender combination, we generated 50 unique prompts, each producing two images, totaling 200 samples per skin tone per model. Samples are annotated following the same procedure as in Sec. \ref{sec:dataset}.

\subsection{Results and Observations}

\begin{figure*}[bt]
  \centering
  \begin{subfigure}{0.32\linewidth}
    \includegraphics[width=\linewidth]{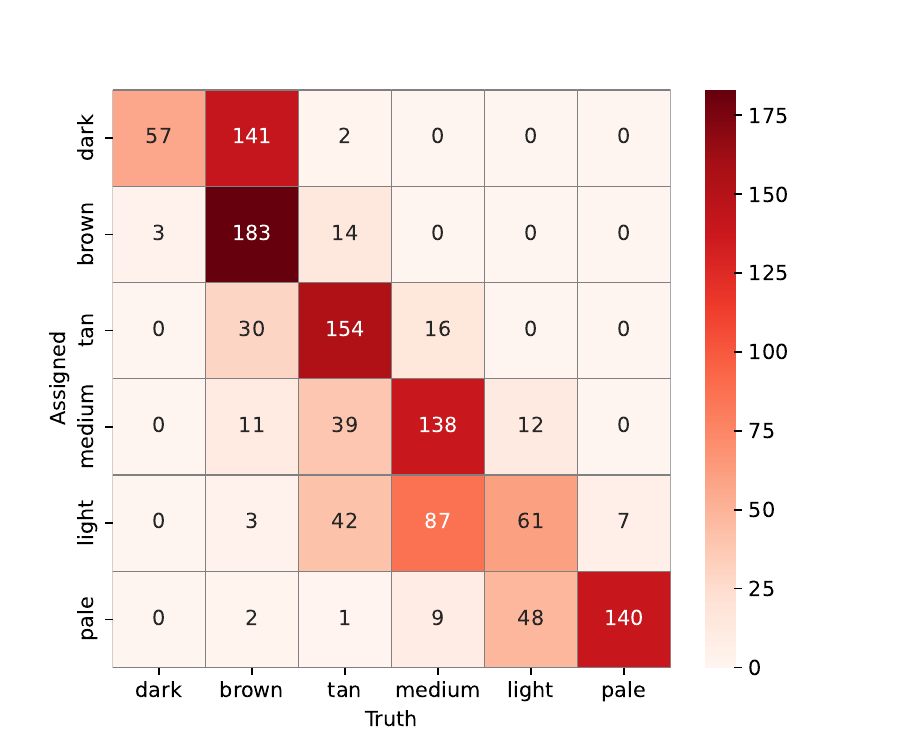}
    \caption{Stable Diffusion XL}
    \label{fig:heat-sdxl}
  \end{subfigure}
  \hfill
  \begin{subfigure}{0.32\linewidth}
    \includegraphics[width=\linewidth]{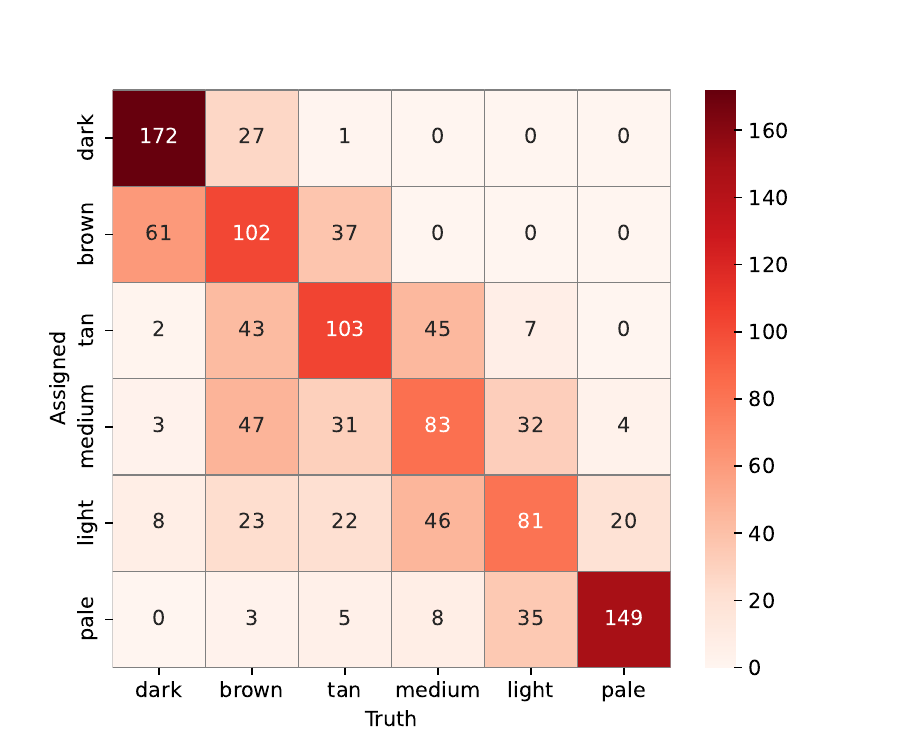}
    \caption{Stable Diffusion 3 Large-Turbo}
    \label{fig:heat-sd3}
  \end{subfigure}
  \hfill
  \begin{subfigure}{0.32\linewidth}
    \includegraphics[width=\linewidth]{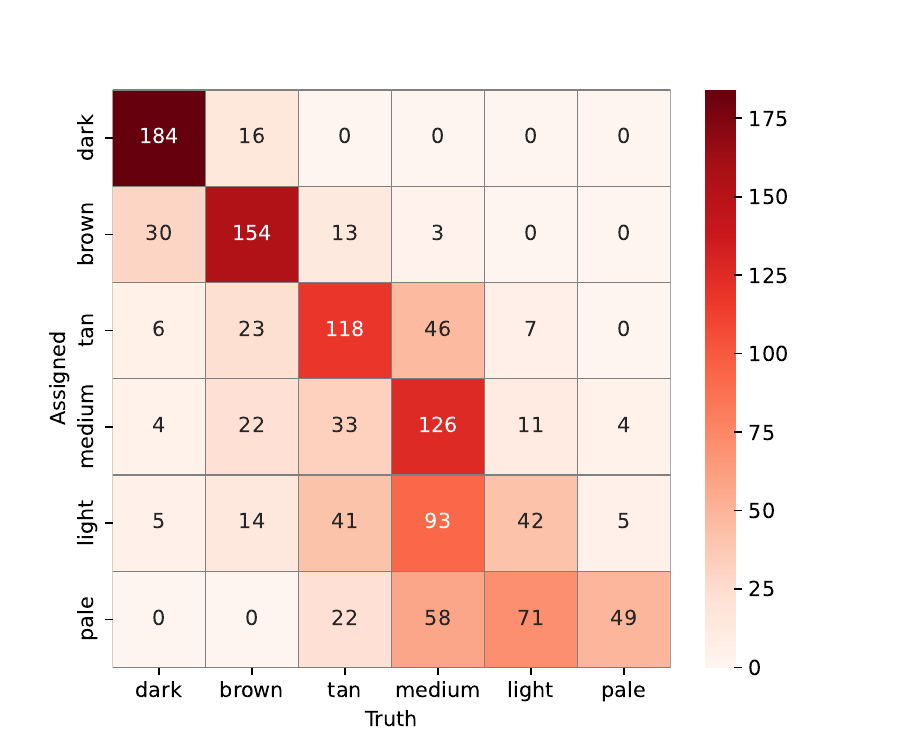}
    \caption{FLUX.1-dev}
    \label{fig:heat-flux}
  \end{subfigure}
  \caption{Confusion matrices for different image generation models. The Y-axis represents the skin tone that the models are supposed to generate according to the text prompt, while the X-axis represents the skin tone actually generated by the models.}  
  \label{fig:heat-gen}
  \vspace{-12pt}
\end{figure*}

Fig. \ref{fig:heat-gen} illustrates the gap between the expected and generated skin tones across models. Unlike LMMs, which perform similarly in recognition, generative models vary in performance: SDXL struggles with dark skin and often confuses light skin with deeper ones; SD3 demonstrates limited ability in distinguishing mid-range skin tones; while Flux underperforms with the lighter tones. \par

\begin{figure}[bt]
  \centering
  \begin{subfigure}{0.49\linewidth}
    \includegraphics[width=\linewidth]{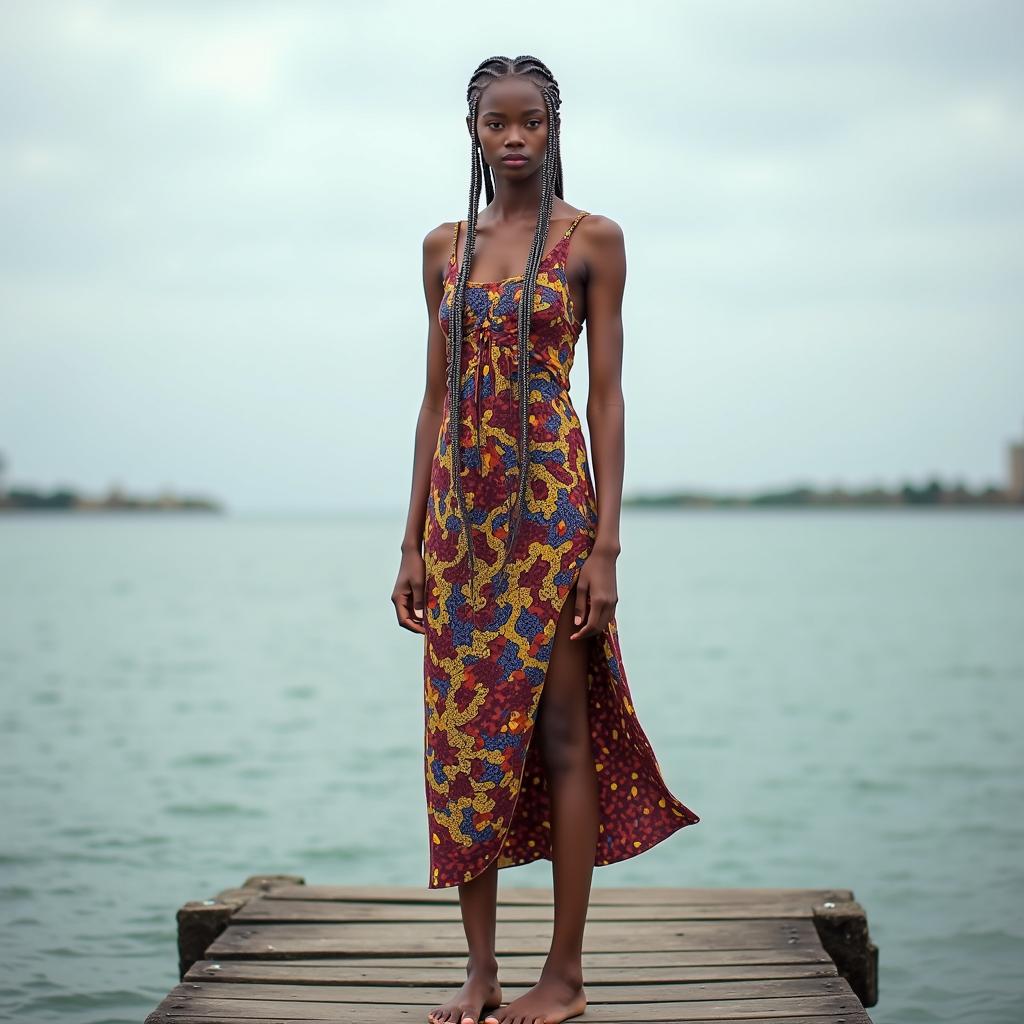}
    \caption{Prompt: \textit{A full-body image of a woman with pale skin, her braided hair adorned with metallic beads, wearing a bold patterned dress, standing barefoot on a pier overlooking calm water.}}    
    \label{fig:pale-0}
  \end{subfigure}
  \hfill
  \begin{subfigure}{0.49\linewidth}
    \includegraphics[width=\linewidth]{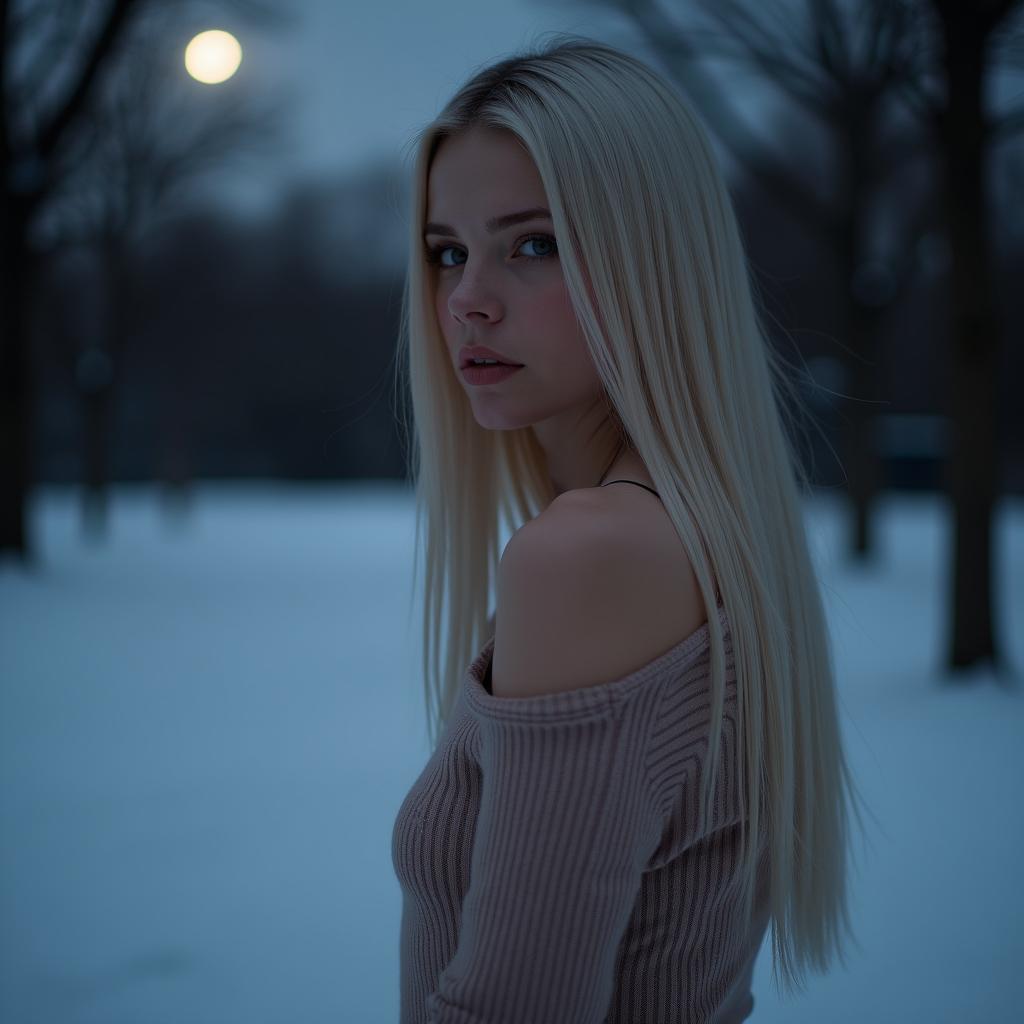}
    \caption{Prompt: \textit{A side-profile portrait of a woman with pale skin, her sleek straight hair glowing softly, illuminated by moonlight, standing barefoot in a snowy park.}}
    \vspace{0.95em}
    \label{fig:pale-1}
  \end{subfigure}    
  \caption{Samples generated with FLUX.1-dev with explicit ``pale skin'' description in their prompts.}  
  \label{fig:gen-bias}
  \vspace{-24pt}
\end{figure}

Beyond overall trends, we observe that generated skin tones are often influenced by other factors in the prompt beyond the skin tone description. For example, with FLUX.1-dev, when the prompt includes a description of braided hair, the generated skin tone tends to be deeper, even if pale is the only specified skin tone descriptor (Fig. \ref{fig:pale-0}). In contrast, when prompts include imagery like snow or nighttime (Fig. \ref{fig:pale-1}), the success rate for generating pale skin increases to 82.5\% (33 out of 40 cases), significantly higher than the overall rate of 24.5\% (49 out of 200 cases). This reflects biases introduced during the training of generative models: individuals with braided hairstyles are more frequently associated with deeper skin tones, while those with pale skin are linked to cold environments. These inherent biases cannot be mitigated through text prompts alone, highlighting the need for a true skin tone dataset that decouples skin tone from other attributes to ensure fairness in image generation models.

\section{Recognition with \dataset}
\label{sec:baseline}

To assess whether \dataset~provides a reliable foundation for training skin tone recognition models and to rule out the possibility that LMMs’ poor performance in Sec.\ref{sec:vlm} stems from benchmark quality issues, we train a baseline recognition model on ~\dataset. The model is built on an EfficientNet-B1~\cite{tan2019efficientnet} model pretrained on ImageNet, with the final layer replaced by a 1280$\times$6 linear layer. \par 
Unlike conventional classification tasks with mutually independent categories, skin tone recognition involves correlated labels. Thus, in addition to accuracy, it is also essential to minimize the deviation of misclassified predictions from their true labels. \par 
To address this feature, we introduce a \textbf{Weighted Cross Entropy Loss}: For a conventional classification task with $N$ samples and $C$ classes, let the logits of the $n$-th $(n=1,2,..N)$ sample be denoted as $z_{n,j}$ $(j\in\{1,2,..C\})$, and the ground-truth label of the $n$-th sample denoted as $y_n$, the softmax probabilities for the $n$-th sample are computed as:
\begin{equation}
p_{n,j} = \frac{\exp(z_{n,j})}{\sum_{k=1}^{C} \exp(z_{n,k})}
\label{eq:softmax}
\end{equation}
\vspace{-4pt}
and the standard cross entropy loss is then defined as:
\begin{equation}
\begin{aligned}
 \mathcal{L}_{\text{CE}} & = -\frac{1}{N} \sum_{n=1}^{N} \log\bigl(p_{n,y_n}\bigr) \\
& = -\frac{1}{N} \sum_{n=1}^{N} \log \left(\frac{\exp(z_{n,y_n})}{\sum_{j=1}^{C}\exp(z_{n,j})}\right)
\end{aligned}
\label{eq:ce}
\end{equation}
In our task, to penalize predictions that deviate significantly from the true labels, we weight the softmax probabilities based on the distance between the predicted and truth labels. Specifically, for sample $n$ and class $j$, we have
\begin{equation}
w_{n,j} = \lambda e^{|y_n-j|}
\label{eq:weight}
\end{equation}
The standard softmax probabilities $p_{n,j}$ are adjusted by multiplication with the weights $w_{n,j}$, followed by re-normalization to ensure a valid probability distribution as:
\begin{equation}
\tilde{p}_{n,j} = \frac{p_{n,j} \cdot w_{n,j}}{\sum_{k=1}^{C} p_{n,k} \cdot w_{n,k}}.
\label{eq:newprob}
\end{equation}
Finally, the Weighted Cross Entropy Loss is calculated over the weighted probabilities:
\begin{equation}
\mathcal{L}_{\text{WCE}} = -\frac{1}{N} \sum_{n=1}^{N} \log\bigl(\tilde{p}_{n,y_n}\bigr).
\label{eq:wcel}
\end{equation}
\vspace{-4pt}

The dataset is divided into 80\% for training, 10\% for validation, and 10\% for testing. The model is trained for 10 epochs with a batch size of 32 using the AdamW optimizer and a learning rate of $1e^{-4}$. The final model achieves an accuracy of $74.18\%$, surpassing LMMs by over 20\%. Moreover, only 2.16\% of the predictions deviate by more than one level from the true label, as illustrated in Fig. \ref{fig:heat-trueskin}. To further evaluate the model's generalization ability, we conducted a zero-shot evaluation on Fitzpatrick17k~\cite{fit17k} dataset\footnote{As discussed in Sec. \ref{sec:dataset}, the six-point Fitzpatrick scale used by Fitzpatrick17k does not perfectly align with \dataset’s labels.}. Our model achieved 30.61\% accuracy and 78.85\% within one level of deviation, outperforming the neural network-based classifier proposed in the paper, which incorporated dataset-specific priors and reached 26.72\% and 60.34\% respectively. \par

These results confirm that the performance of LMMs is not misestimated due to benchmark quality and demonstrates that ~\dataset~serves as a strong foundation for training high-performance true skin tone recognition models.

\section{Fine-tuning with \dataset}
\label{sec:finetune}
To evaluate whether \dataset~can mitigate bias in skin tone synthesis in image generation models, we fine-tune a representative generative model on \dataset~and analyze performance changes throughout training. For real images, prompts are generated by JoyCaption~\cite{joycaption2025} with explicit descriptions of skin tone. For generated images, the prompts correspond to those used during their generation. It is worth noting that iteratively training the generation model on a curated subset of generated images is a common and effective approach to improve model quality. Specifically, we trained a LoRA~\cite{hu2022lora} model with rank $16$ on SDXL, using two A100 GPUs, a batch size of $4$, and a learning rate of $1e^{-4}$. \par

\begin{figure}[ht]
\centering
\begin{tabular}{ccc}
\includegraphics[width=0.135\textwidth]{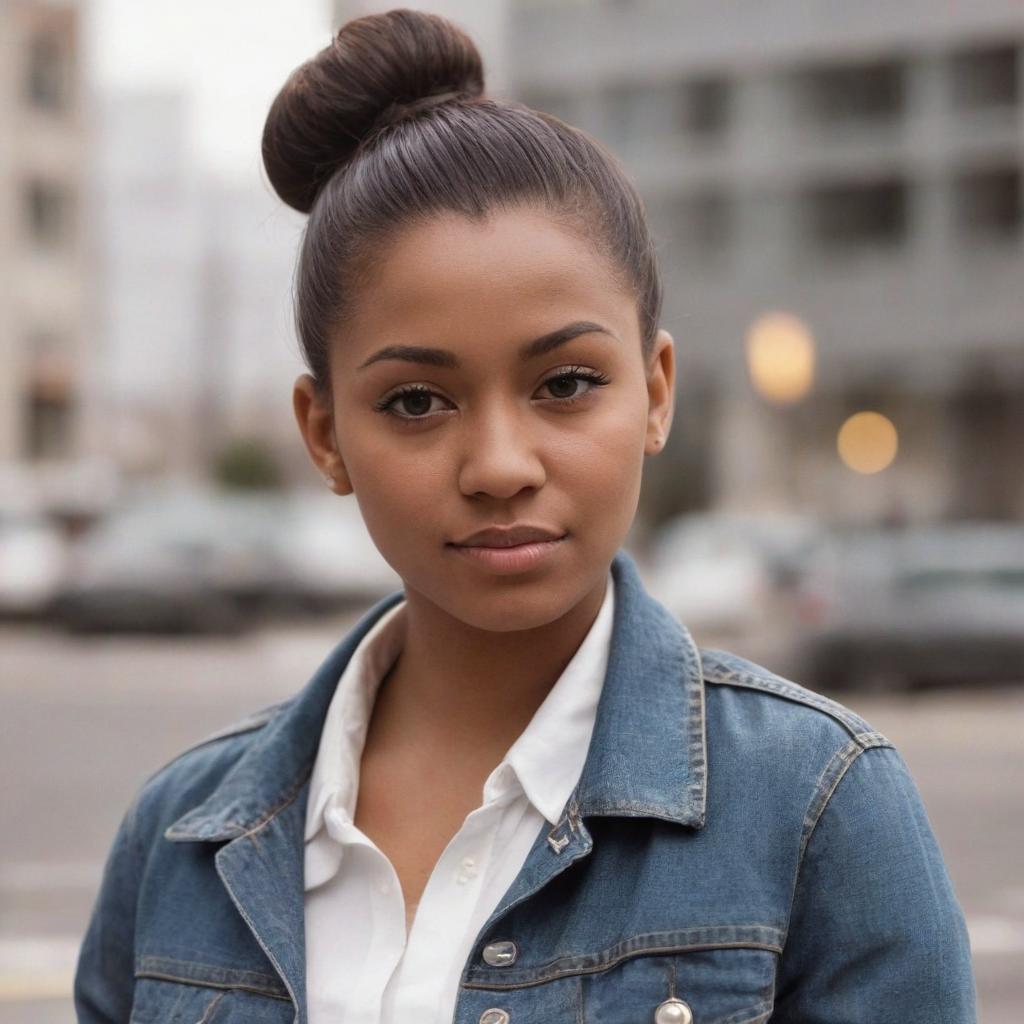} & \includegraphics[width=0.135\textwidth]{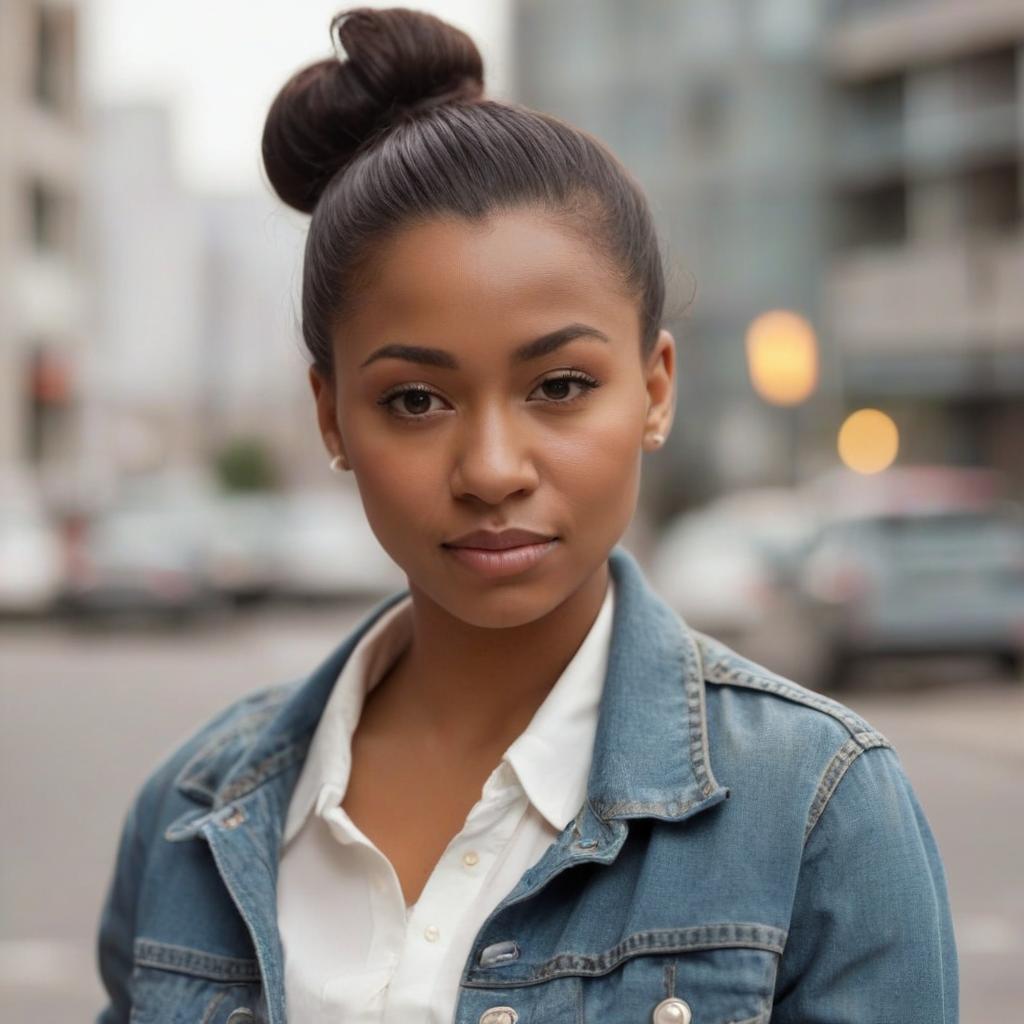} & \includegraphics[width=0.135\textwidth]{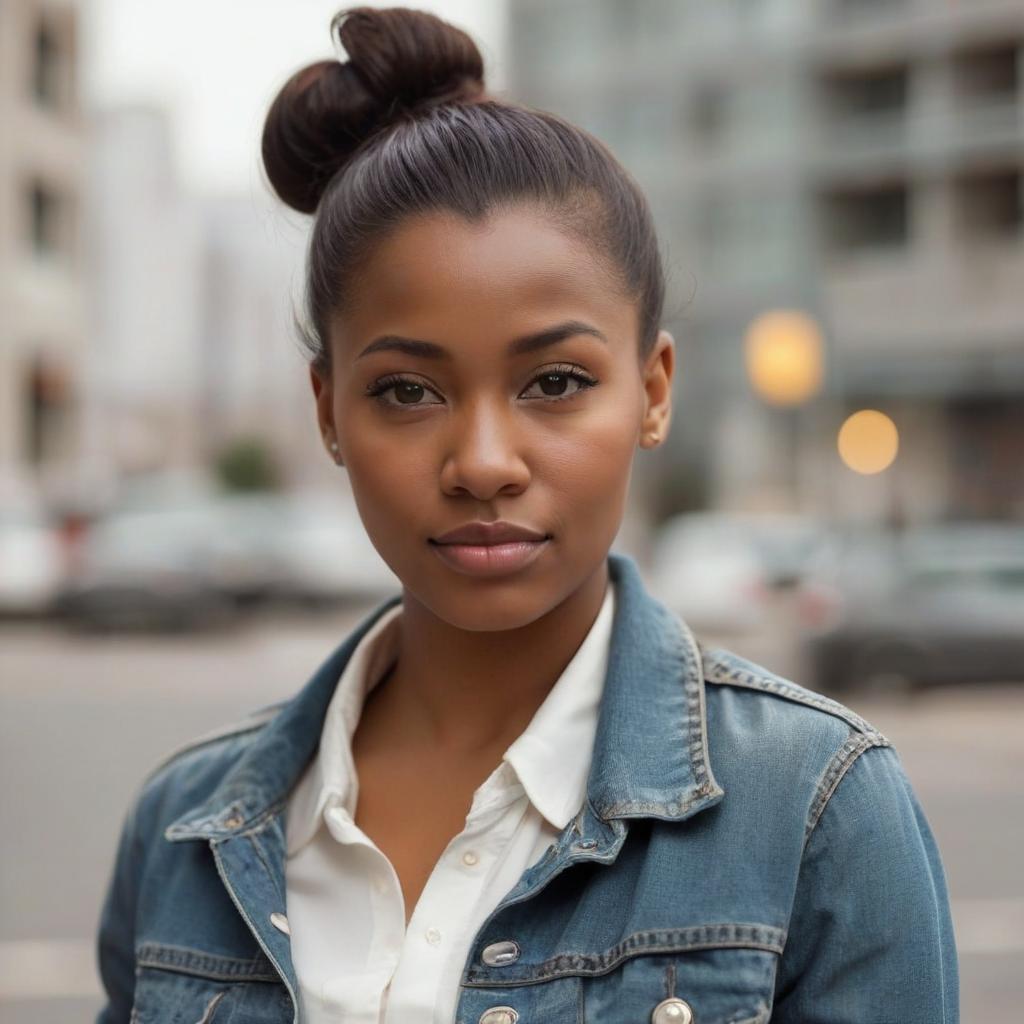} \\ 
\includegraphics[width=0.135\textwidth]{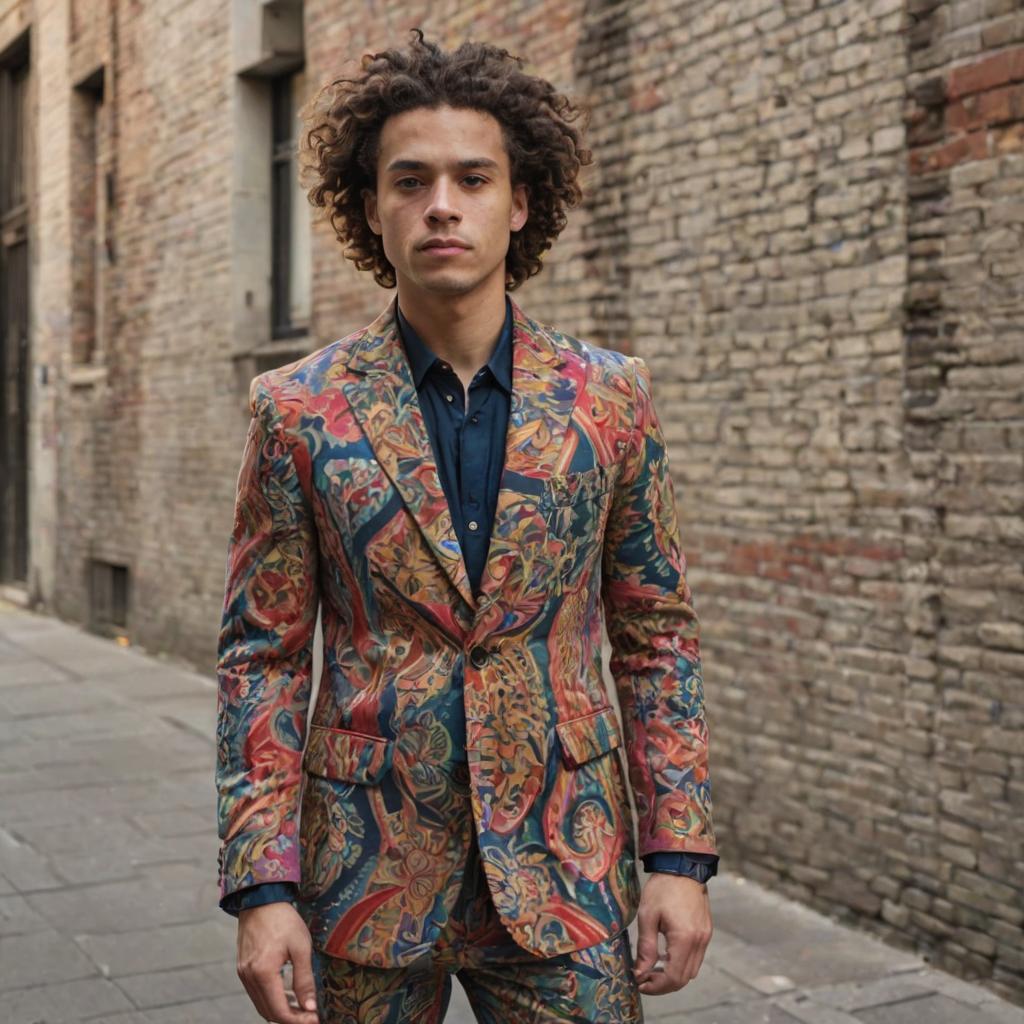} & \includegraphics[width=0.135\textwidth]{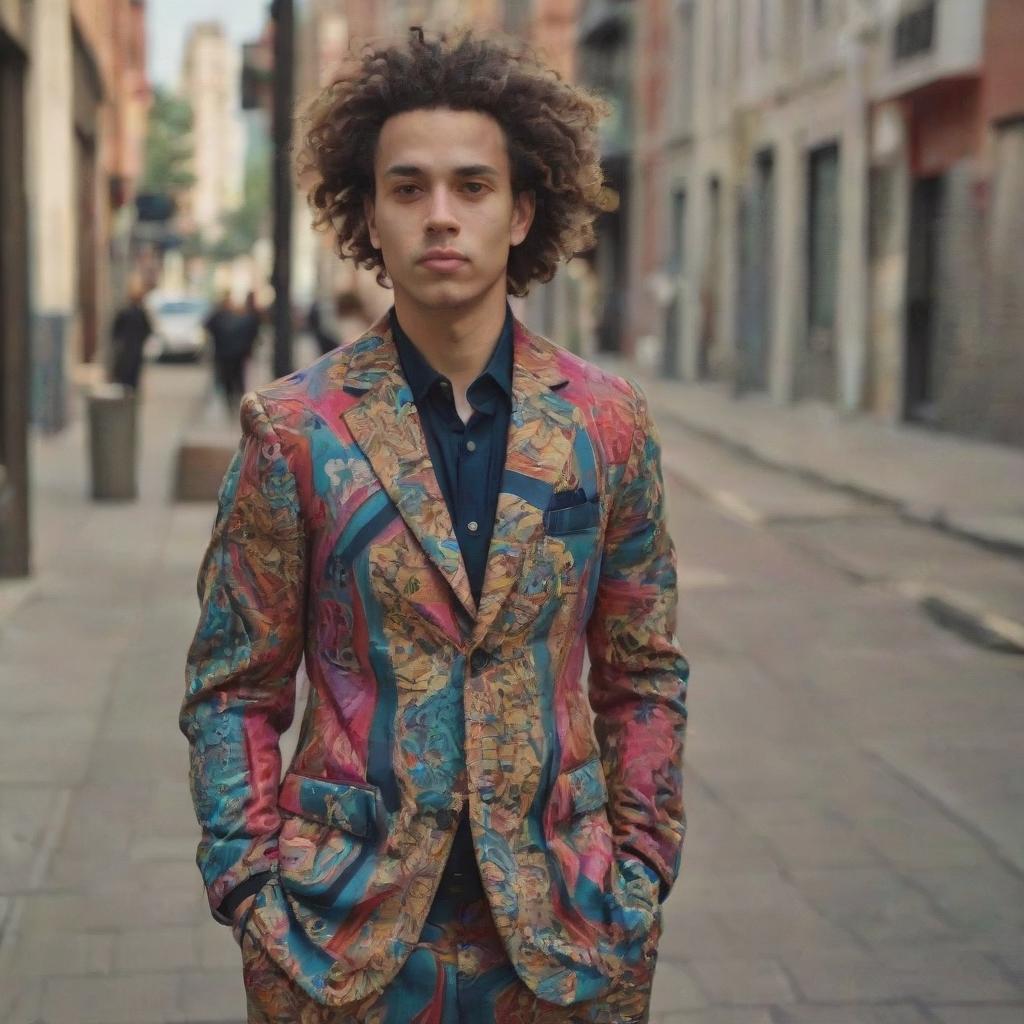} & 
\includegraphics[width=0.135\textwidth]{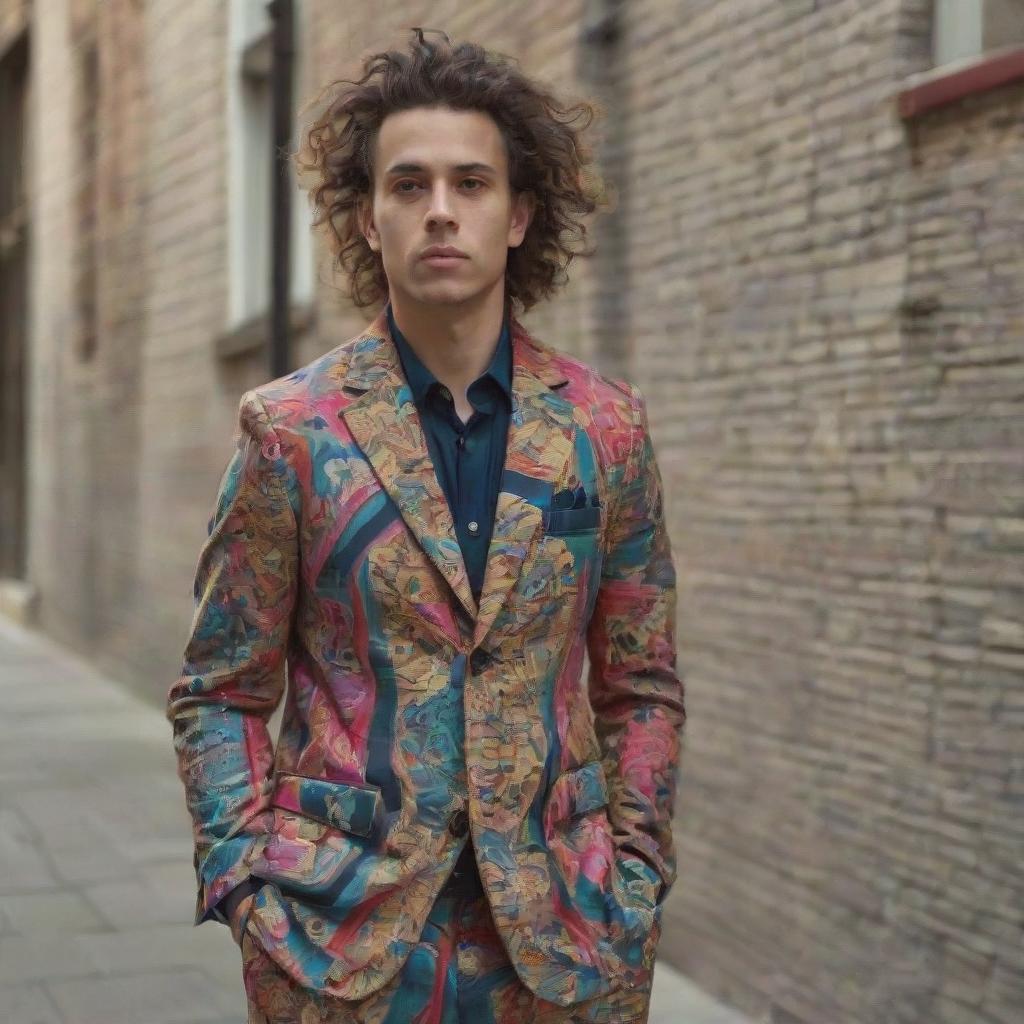} \\
step 0 & step 200 & step 400 \\
\end{tabular}
\caption{Changes in generated skin tones during fine-tuning. Top: \textit{A half-body portrait of a woman with dark skin, wearing a denim jacket over a white shirt, her hair in a sleek bun, with soft natural lighting and a blurred urban background.}\\Bottom: \textit{A full-body image of a man with light skin, his voluminous hair catching the light, wearing a bold patterned blazer, standing confidently in a vibrant urban setting.}}
\label{fig:finetune}
\end{figure}

Fig. \ref{fig:finetune} demonstrates the changes during training concerning the two issues of SDXL discussed in Sec. \ref{sec:diffusion}. Each row of images is generated using the same text prompt and random seed. The top row shows how fine-tuning reduces SDXL's bias toward brown when dark skin is expected, with the generated tone gradually darkening. The bottom row illustrates the influence of other elements on skin tone. Despite specifying light skin, the base model generates a deeper tone due to inherent bias. Fine-tuning gradually reduces this bias, bringing the output tone closer to the target. Table. \ref{tab:finetune-stat} quantifies the model's improvement during training, showing increased accuracy and reduced deviation in generated skin tones.

\begin{table}[tb]
\caption{Accuracy and mean squared error (MSE) of the generated vs. target skin tones. Row ``SDXL (step 0)'' refers to the original SDXL model before fine-tuning.}
    \label{tab:finetune-stat}
    \centering
    \begin{tabular}{lcc}
        \toprule
         Model & Accuracy & MSE \\
         \midrule
         FlUX.1-dev & 56.08\% & 1.1208 \\
         SD3 Large Turbo & 57.50\% & 0.9967 \\
         SDXL (Step 0) & 61.08\% & 0.6008 \\
         SDXL (Step 400) & 63.08\% & 0.5142 \\
         SDXL (Step 800) & 64.75\% & 0.4800 \\
         \bottomrule
    \end{tabular}
\vspace{-10pt}
\end{table}
\section{Conclusion}
\label{sec:conclusion}

In this paper, we introduce \dataset, a diverse and carefully annotated dataset designed for true skin tone recognition and generation tasks. Through comprehensive benchmarking, we reveal significant biases in existing LMMs and generative models, highlighting their limitations in accurately predicting and synthesizing specified skin tones. Training and fine-tuning models using \dataset~demonstrate substantial performance improvements, confirming its effectiveness in addressing dataset biases and supporting more accurate and fair skin tone representation across recognition and generation tasks. \par

Although TrueSkin addresses several key limitations of existing datasets, it still has constraints such as (1) coarse-grained labeling that cannot precisely capture subtle variations between yellowish and reddish skin tones, and (2) residual annotation subjectivity despite mitigation efforts. To further advance this field, future work will focus on expanding the dataset’s scale and diversity, developing automated or semi-automated annotation tools leveraging multimodal or self-supervised models to reduce human subjectivity, and integrating causal modeling techniques to disentangle skin tone from environmental factors.

\bibliographystyle{IEEEtran}
\bibliography{main}

\end{document}